\documentclass{article} %
\usepackage{iclr2026_conference,times}

\usepackage{amsmath,amsfonts,bm}

\def\eqref#1{equation~\ref{#1}}

\def\1{\bm{1}}

\DeclareMathAlphabet{\mathsfit}{\encodingdefault}{\sfdefault}{m}{sl}
\SetMathAlphabet{\mathsfit}{bold}{\encodingdefault}{\sfdefault}{bx}{n}

\DeclareMathOperator*{\argmax}{arg\,max}

\usepackage{hyperref}
\usepackage{url}
\usepackage{cleveref}
\usepackage{amsmath, amssymb, amsfonts}
\usepackage{enumitem}
\usepackage{tabularx}
\usepackage{tikz,nicematrix}
\usepackage[ruled, linesnumbered, nofillcomment,vlined]{algorithm2e}
\usepackage{listing}
\usepackage{threeparttable}
\usepackage{graphicx}
\usepackage{siunitx}
\usepackage{array,float}
\usepackage{booktabs}
\usepackage{xcolor}
\usepackage{wrapfig}
\usepackage{adjustbox}
\usepackage{balance}
\usepackage{natbib}
\usepackage{calc}
\usepackage{caption, subcaption}

\usepackage{pifont}%
\newcommand{\cmark}{\ding{51}}%
\newcommand{\xmark}{\ding{55}}%

\crefname{algocf}{alg.}{algs.}
\Crefname{algocf}{Algorithm}{Algorithms}

\title{NFPO: Stabilized Policy Optimization of Normalizing Flow for Robotic Policy Learning}

\author{Diyuan Shi$^{1,2,*}$, Yiqi Tang$^{2,*}$, Zifeng Zhuang$^{2}$, Donglin Wang$^{2,\dagger}$ \\ {\small $^{1}$Zhejiang University \, $^{2}$Westlake University \, $^{*}$Equal Contribution \,  $^{\dagger}$Corresponding Authors }   }

\iclrfinalcopy %
\begin{document}

\maketitle

\begin{abstract}
Deep Reinforcement Learning (DRL) has experienced significant advancements in recent years and has been widely used in many fields. In DRL-based robotic policy learning, however, current \textit{de facto} policy parameterization is still multivariate Gaussian (with diagonal covariance matrix), which lacks the ability to model multi-modal distribution. In this work, we explore the adoption of a modern network architecture, i.e. Normalizing Flow (NF) as the policy parameterization for its ability of multi-modal modeling,  closed form of log probability and low computation and memory overhead. However, naively training NF in online Reinforcement Learning (RL) usually leads to training instability. We provide a detailed analysis for this phenomenon and successfully address it via simple but effective technique. With extensive experiments in multiple simulation environments, we show our method, NFPO could obtain robust and strong performance in widely used robotic learning tasks and successfully transfer into real-world robots.
\end{abstract}

\section{Introduction}
Deep Reinforcement Learning (DRL), as a machine learning field studying sequential decision making, has received tremendous research interest and advancements in recent years \citep{mnihHumanlevelControlDeep2015,ouyangTrainingLanguageModels2022a,deepseek-aiDeepSeekR1IncentivizingReasoning2025}. With reduced Sim-to-Real gap \citep{margolisRapidLocomotionReinforcement2022c,fuDeepWholeBodyControl2022}, nowadays   the policies trained in parallel simulation environments by RL could be transferred with unprecedented easiness to real world robotic systems. This avoids the need to manually collect real data which is highly expensive and cumbersome, significantly speeding up the researches and developments. Through this way, many policies trained by DRL have been successfully deployed  and enabled impressive accomplishments in generating smooth, autonomous motions on real-world bipedal, quadruped and humanoid robots \citep{liRobustVersatileBipedal2023,leeLearningQuadrupedalLocomotion2020a,zeTWISTTeleoperatedWholeBody2025,margolisRapidLocomotionReinforcement2022b,radosavovicRealworldHumanoidLocomotion2024,zhangFALCONLearningForceAdaptive2025,shaoLangWBCLanguagedirectedHumanoid2025}.

However, current \textit{de facto} policy parameterization in DRL robotic policy learning is still multivariate Gaussian with diagonal covariance matrix which is known to have poor ability to model mutli-modal distributions. This is in stark contrast to another paradigm, Supervised-Learning-Based robotic policy learning (a.k.a Behavior Cloning where a policy is trained to mimic pre-collected behavior dataset) where modern policy networks like Diffusion Policy \citep{chiDiffusionPolicyVisuomotor2023,liaoBeyondMimicMotionTracking2025} and Transformers \citep{RT-1,RT-2} are in wide adoption.

While many recent works have studied the integration of diffusion-based policies into Online Reinforcement Learning paradigm, most of them are built on off-policy methods (like Soft Actor Critic (SAC) \citep{haarnojaSoftActorCriticOffPolicy2018}) for stronger sample efficiency. While in robotic learning, simulation environments offer massive samples  with \textit{trivial  cost} and it's 1) computation efficiency, 2) memory consumption and 3) robustness towards multiple simulator and reward function settings  are of more importance. Under this setting, on-policy methods like Proximal Policy Optimization (PPO) \citep{schulmanTrustRegionPolicy2017,schulmanProximalPolicyOptimization2017} are more favored and have delivered countless successes. 
Unfortunately, it remains unclear how to integrate modern multi-modal-modeling networks into Policy Optimization's paradigm,  and train control policy purely from scratch\footnote{There also exist methods that perform purely offline or offline-to-online finetuning of diffusion or Transformer policies via RL for robots. But we focus on from-scratch DRL learning where no dataset is present.}. A table comparison between our method and related works could be found in \Cref{tbl:algo_compare}.

In this work, we aim to bridge this gap by designing a new method which is 1) computation and memory efficient, 2) robust towards multiple simulator and reward function settings and 3) simple with few code changes to current training pipeline. We choose Normalizing Flow (NF) as it naturally fits all above requirements. However, naively combining NF with Policy Optimization would cause severe training and numerical instability. We provide detailed analysis and show how to address it with simple but effective techniques. In summary, our contributions are:
\begin{enumerate}[leftmargin=*]
   \item Aiming at robotic multi-modal policy learning, we integrate NF into PPO, analyze the reasons of its training instability and propose methods to address it. 
   \item We extensively test our method in multiple widely-used simulation environments (IsaacGym, Mujoco-playground and IsaacLab) and demonstrate our method (with the same set of configurations) could obtain competitive performance compared to state-of-the-art Gaussian PPO implementation.
   \item We successfully transfer the policy trained with NFPO to real-world robots to show it could perform various tasks like locomotion and motion tracking.
\end{enumerate}
\vspace{-8pt}
\begin{table}[t]
    \centering
      \caption{Comparison of NFPO to related methods. Detailed explanations of how we choose \cmark or \xmark \; and the references are provided in \Cref{app:table_details}.}
      \label{tbl:algo_compare}
   \begin{threeparttable}
      \newcolumntype{C}[1]{>{\centering\arraybackslash}m{#1}}

        \begin{NiceTabular}{|C{3.2cm}|C{1.5cm}|C{1.5cm}|C{2.2cm}|C{2.8cm}|}[hvlines]
            \CodeBefore
                \rowcolor{lightgray}{1-1}
            \Body
            Algorithm & Multi-modality & On-policy & Memory \& Computation&  Real World Deployment \\
            FastTD3  & \xmark & \xmark & \xmark &  \cmark \\
            Meow  & \cmark & \xmark & \cmark &  \xmark \\
            MaxEntDP  & \cmark & \xmark & \xmark &  \xmark \\
            GenPO  & \cmark & \cmark & \xmark &  \xmark \\
            FPO$^\alpha$  & \cmark & \cmark & \cmark &  \xmark \\
            \textbf{Ours (NFPO)} & \cmark & \cmark & \cmark &  \cmark \\
            
        \end{NiceTabular}
         \begin{tablenotes}
            \footnotesize
            \item[$\alpha$] This work is finished concurrently to FPO. After that, a recent work, FPO++ \citep{yi2026FlowPolicyGradientsRobotControl} could be successfully deployed onto robotic plarform.
        \end{tablenotes}
    \end{threeparttable}
    \vspace{-14pt}
\end{table}

\section{Related Works}
\textbf{Deep Reinforcement Learning}. DRL has experienced tremendous advancements in recent years. From on-policy methods like TRPO \citep{schulmanTrustRegionPolicy2017}, PPO \citep{schulmanProximalPolicyOptimization2017} to off-policy methods like DQN \citep{mnihHumanlevelControlDeep2015}, DDQN \citep{vanhasseltDeepReinforcementLearning2015}, TD3 \citep{fujimotoAddressingFunctionApproximation2018a} and SAC \citep{haarnojaSoftActorCriticOffPolicy2018}, a main research direction is to improve the \textit{sample-efficiency} which measures how many samples a method needs to achieve certain return, as humans could learn very efficiently with few interactions. By leveraging learnable dynamics and reward models, Model-based RL like Dreamer \citep{hafnerMasteringDiverseControl2025} and TDMPC \citep{hansenTDMPC2ScalableRobust2024} further increases the sample-efficiency compared to their model-free alternatives. Inspired by recent innovations in neural network architecture like Diffusion Model, Transformers and Normalizing Flows, another line of works targeting on combining these powerful networks into DRL's framework has also emerged as in \citet{chaoMaximumEntropyReinforcement2024,dingGenPOGenerativeDiffusion2025,dongMaximumEntropyReinforcement2025}. In this work, we try to integrate NFs into on-policy PPO's pipeline as the latter is widely used in robotic policy learning.

\textbf{Normalizing Flow}. Normalizing Flow is a kind of generative models, featured by bijective mapping and closed calculation of log probability, compared to other generative methods like GAN \citep{goodfellow2014GenerativeAdversarialNetworks}, Diffusion Models \citep{hoDenoisingDiffusionProbabilistic2020} and Score-based  Models \citep{songScoreBasedGenerativeModeling2020}. Since the early works like \citet{dinhNICENonlinearIndependent2015,dinhDensityEstimationUsing2017}, NFs have also gained significant improvement. Especially in \citet{zhaiNormalizingFlowsAre2025}, the flexibility and generation quality of NFs have gained much improvement  by using attention techniques. In DRL, the efficient and accurate calculation of log probability has made NFs an appealing policy parameterization and some works have explored the combination of NFs and DRL as in \citep{chaoMaximumEntropyReinforcement2024,ghugareNormalizingFlowsAre2025}. However, to the best of our knowledge, we are the first to integrate NFs into  on-policy RL setting.

\textbf{Robotic Policy Learning}. Thanks to the rapid advancements of GPU-based parallel simulation environments and reduced Sim-to-Real gap \citep{kumarRmaRapidMotor2021b,fuDeepWholeBodyControl2022}, Robotic Policy Learning has garnered great progress. From bipedal, quadruped robots with locomotion skills like parkour \citep{eth_dog}, fast running \citep{margolisRapidLocomotionReinforcement2022a} to humanoid robots where teleoperation \citep{zeTWISTTeleoperatedWholeBody2025,he2024OmniH2OUniversalDexterousHumantoHumanoidWholeBodyTeleoperationLearning} and motion tracking \citep{liaoBeyondMimicMotionTracking2025} have enabled  dancing, kicking and somersault, many agile and stable behaviors have been learned by DRL pipeline. Alternatively, in Supervised-Learning-based robotic policy learning, modern network architecture like Transformers and Diffusion Models have shown great advantages for their expressiveness, like in \citet{chiDiffusionPolicyVisuomotor2023,renDiffusionPolicyPolicy2024,RT-2,octo_2023}. In this work, we try to explore the integration of NFs as policy parameterizations in DRL paradigm.

\section{Background}
\textbf{Reinforcement Learning}. Reinforcement Learning (RL) aims to solve sequential decision-making problem which is formulated as a Markovian Decision Process (MDP): $\mathcal{M} = \{\mathcal{S}, \mathcal{A}, \mathcal{R}, P, \gamma, \mathcal{S}_0 \}$ where $\mathcal{S}$ is the set of all \textit{states},  $\mathcal{A}$ is the set of all \textit{actions}, $\mathcal{R}: (s_t, a_t) \rightarrow \mathbb{R}$ is the reward function that gives a scalar value for given state action pair, $P: s_{t+1} \sim P(s_t, a_t)$ is the transition model that gives the next state given current state action, $\gamma$ is the discount factor and $\mathcal{S}_0$ is the distribution of initial states. RL is to find a policy function $\pi$ that maximizes the expected return in $\mathcal{M}$:
\begin{equation}
    \pi = \argmax\limits_{\pi \sim \Pi}  \mathop{\mathbb{E}}\left[ \sum\limits_{t=0}^{\infty} \gamma^t \mathcal{R}(s_t, a_t) \right] 
\end{equation}
\textbf{Proximal Policy Optimization}. As a notable variant of on-policy RL, Proximal Policy Optimization (PPO) \citep{schulmanProximalPolicyOptimization2017} performs policy optimization via advantage-based clip updates:
\begin{equation}
   \pi_{\text{new}} = \underset{\theta}{\argmax} \underset{a \sim \pi_{\text{old}}, \\ s \sim P(s, a)}{\mathbb{E}}\Bigl[\min \bigl( r(\theta)\tilde{A}(s,a), \text{clip} (r(\theta), 1-\epsilon, 1+\epsilon) \tilde{A}(s,a) \bigr)\Bigr]
\end{equation}
where $r(\theta) = \frac{\pi_{\theta}(a \vert s)}{\pi_{\text{old}}(a \vert s)}$ is the ratio of action likelihood and $\tilde{A}(s,a)$ is the estimated advantages (e.g., Generalized Advantage Estimation (GAE) as in \citet{schulmanHighDimensionalContinuousControl2018a}). %
As it requires log probability of action given state, the common policy parameterization is Gaussian with learnable mean and diagonal covariance matrix: \textcolor{black}{$\pi(s) = \mathcal{N}(\mu_\theta(s), \text{diag}(\sigma_\theta(s)))$}.

\textbf{Normalizing Flow}. Normalizing Flow (NF) is a bijective mapping with learnable components $f_\theta: \mathbb{R}^D \rightarrow \mathbb{R}^D$ between data distribution $p(x)$ and prior distribution $q(z)$. It's designed such that the inverse $f^{-1}_\theta$ and the determinant of its Jacobian $\vert \frac{df_\theta(x)}{dx} \vert$ is in closed form and could be efficiently computed. The prior distribution is chosen as simple ones like Normal distribution. Then the data density could be expressed as:
\begin{equation}
  p(x) = q(f_\theta(x))  \bigg| \frac{df_\theta(x)}{dx} \bigg|
\end{equation}

Then we could use Maximum Likelihood Estimation (MLE) based training objectives $\theta = \argmax_\theta \mathbb{E}_{x} [\log q(f_\theta(x)) + \log \vert \frac{df_\theta(x)}{dx} \vert]$, and perform \textit{sampling} via $z \sim q(z); \, x = f^{-1}_\theta(z)$ and \textit{inference} via $x \sim p(x); \, z = f_\theta(x)$.

\textbf{RealNVP}. An important NF variant is RealNVP \citep{dinhDensityEstimationUsing2017}  with \textit{coupling layers}:
\begin{align}
   f_\theta(x)_{d} &= x_{d} \\
   f_\theta(x)_{\backslash d} &= x_{\backslash d} \odot \exp(s_\theta(x_{d})) + t_\theta(x_{d})
\end{align}
where $d$ is a set containing certain indexes, $\backslash d$  $\Bigl(=\{i \vert 1 \le i \le D, i \in \mathbb{N}^{+} \} \backslash d\Bigr)$ is a set containing remaining indexes, $\odot$ is element-wise production and $s_\theta$, $t_\theta$ are 2 neural network from $\mathbb{R}^{\vert d\vert}$ to $\mathbb{R}^{\vert D-d \vert}$. From above definition, the Jacobian is: $\vert \frac{df_\theta(x)}{dx} \vert = \exp\bigl[\sum s(x_d)\bigr]$. Following the original work, we employ alternating odd-even strategy to partition the index set in which 3 or more stacked layers is needed to allow each dimension in $x$ to influence each other. 

\section{Method}
With above definition of PPO and NFs, we could build an optimization pipeline where NFs are used as policy parameterizations instead of Gaussian. Thanks to its closed form of log probability, the changes mainly involve a new calculation of $\pi(a \vert s)$ and sampling from prior distribution to generate action samples. All other components and formulas could be reused.
However, a naive applying of above methods would likely result in training instability, for the following reasons:

\newlength{\strutheight}
\settoheight{\strutheight}{\strut}
\begin{enumerate}[leftmargin=*]
   \item
   \begin{adjustbox}{valign=T,raise=\strutheight,minipage={.98\linewidth}}
   \begin{wrapfigure}[22]{r}{0.5\textwidth}
      \vspace{-26pt}
   \centering
   \includegraphics[width=0.5\textwidth]{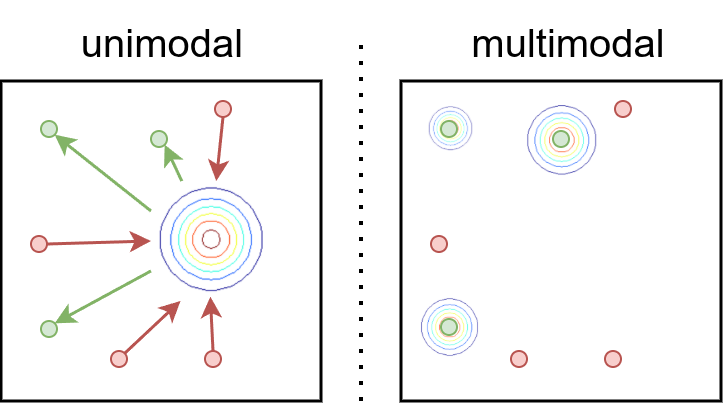}
   \vspace{-15pt}
   \caption{Illustrative diagram of multi-modal overfitting. Green samples are those whose probability needs increase while red samples need decrease. In left figure, the unimodal model cannot fit all these training signals but in the right figure a multimodal model overfits.}
   \label{Fig:motiv_draw}
   \end{wrapfigure}
    \strut{}\textbf{Overfitting}. Expressive multi-modality policies are prone to overfitting than Gaussian-based counterparts. As the modeling ability increases, the network could find a cheap optimization strategy that assigns high log-probability to all points with positive advantages and vice versa. An illustrative diagram is in \Cref{Fig:motiv_draw}.
    \end{adjustbox}
   
   \item
   \begin{adjustbox}{valign=T,raise=\strutheight,minipage={.98\linewidth}}
   \begin{wrapfigure}[16]{r}{0.5\textwidth}
   \centering
   \parbox[c][20pt][c]{.55\linewidth}{}

   \end{wrapfigure}
   \strut{}\textbf{Exponential values}. the $\exp(s_\theta(x_{d}))$ used in RealNVP applies exponential transformations on the output of neural networks. This further increases the tendency of overfitting as simply increase $s_\theta(x_d)$ would increase the training objective significantly.
    \end{adjustbox}
   \item 
   \begin{adjustbox}{valign=T,raise=\strutheight,minipage={.98\linewidth}}
   \begin{wrapfigure}[1]{r}{0.5\textwidth}
   \centering
   \parbox[c][20pt][c]{.55\linewidth}{}

   \end{wrapfigure}
   \strut{}\textbf{Unbounded Output}. Unlike Gaussian whose log-probability is roughly bounded (so long as the standard deviation is not infinitesimal), neural network's output could be unbounded and lead to numerical instability values.
    \end{adjustbox}

\end{enumerate}
To further illustrate above phenomenon, we perform an experiment using UnitreeRLGym's g1 environment. Specifically, we build a 4-layered RealNVP as policy parameterization and integrate it into PPO's training pipeline, the training result is in \Cref{Fig:motiv} under name of \texttt{s\_none}.

\begin{wrapfigure}[16]{r}{0.5\textwidth}
\vspace{-18pt}
\centering
\includegraphics[width=0.5\textwidth]{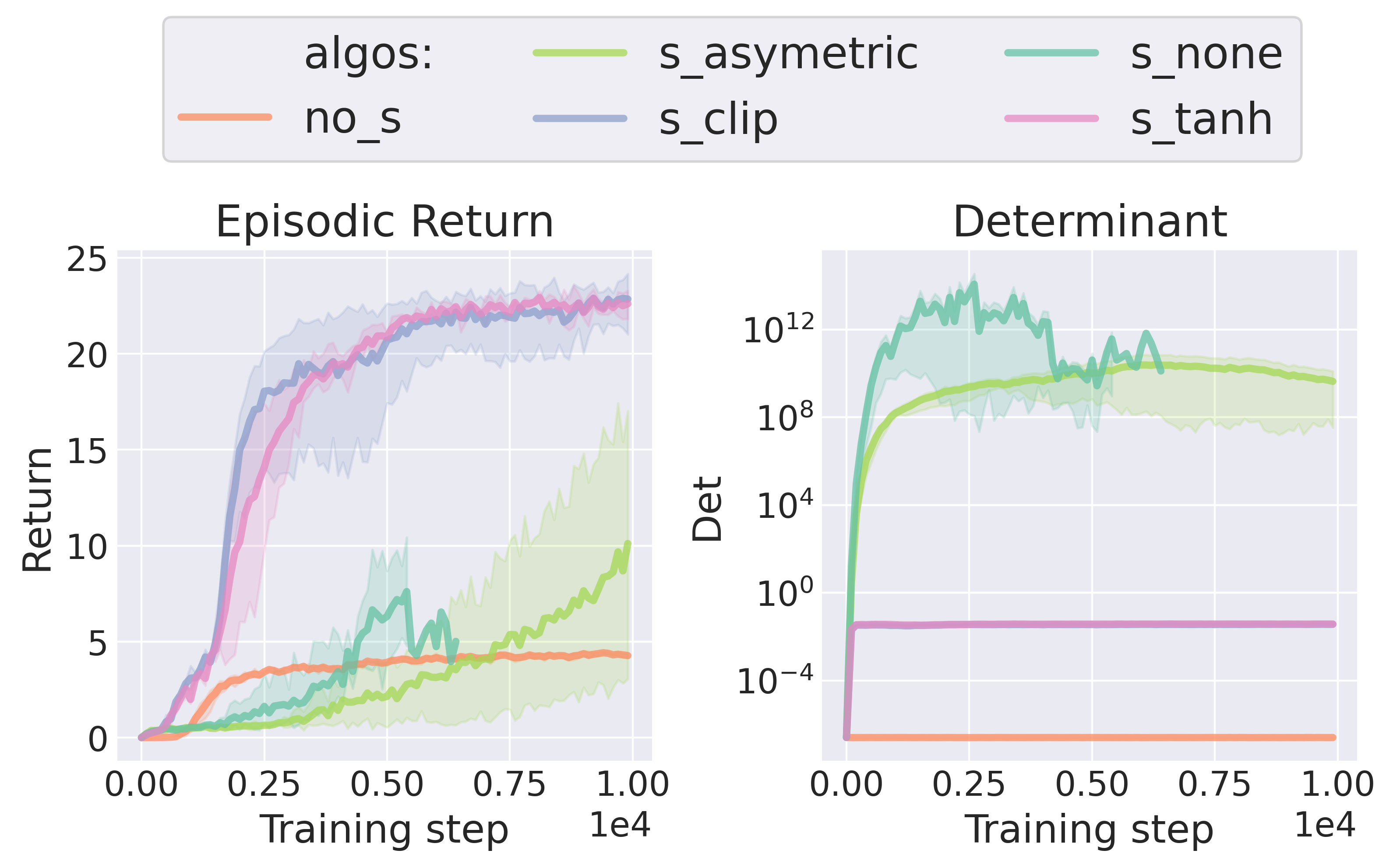}
\vspace{-16pt}
\caption{Training NFs in Unitree Gym's g1. The experiment is in 3 seeds with 95\% confidence interval. \texttt{s\_none} early stopped due to training instability, \textcolor{black}{\texttt{s\_clip} and \texttt{s\_tanh} overlaps in determinant plot.}}
\label{Fig:motiv}
\end{wrapfigure}

From the result, we find the performance of \texttt{s\_none} increases in the first and middle stage of training. However, the determinant of its Jacobian keeps increasing to very large values then it triggers numeric instability and training crashes.

\textbf{Solution}. As observed above, the training instability of NFs under Policy Optimization is mainly caused by the  exponential transformation of unbounded output of $s_\theta(x)$. A simple yet effective technique is to `normalize' the $s_\theta(x)$ output to make it in proper range. In this section, we test various methods for normalizing $s(x)$ to make it in bounded range. In details, we test 1) \texttt{no\_s} where we omit $s(x)$ and only use $t(x)$. This is reported in \citet{chaoMaximumEntropyReinforcement2024} to have sufficient expressiveness in online RL setting, 2) \texttt{s\_clip} where a $\text{clip}(s_\theta(x), -l, l)$ is applied with a hyperparameter $l$, 3) \texttt{s\_tanh} where we use $l\times \text{tanh}(s_\theta(x))$ and 4) \texttt{s\_asymetric} which is reported in \citet{andradeStableTrainingNormalizing2024} as an advanced `normalizing' technique. We train all of these methods (with $l=0.5$) with PPO in Unitree RL Gym's g1 environment, and the result is in \Cref{Fig:motiv}.

From the result, we find \texttt{s\_clip} and \texttt{s\_tanh} are 2 most effective `normalizing' methods and both \texttt{no\_s} and \texttt{s\_none} obtains insufficient performance. While \texttt{s\_asymetric} could obtain performance increase in the ending of training, the determinant of it is still in high scale.  Finally, we choose \texttt{s\_tanh}  for its simplicity and robustness (in \Cref{sect:q1}  we compare \texttt{s\_tanh} and \texttt{s\_clip} in experiments \textcolor{black}{and provide an anlaysis over why tanh may be better than clip in \Cref{app:theory_of_tanh_clip}}).

Finally, we could build a stabilized PPO-NF method which we name as NFPO (\textcolor{black}{for brevity, the conditioned state $s$ is omitted}):
\begin{equation}\label{eq:1}
\begin{aligned}
      \pi_{\text{new}} = \underset{\theta}{\argmax} \; \mathbb{E} &\Bigl[\min \bigl( r(\theta)\tilde{A}(s,a), \text{clip} (r(\theta), 1-\epsilon, 1+\epsilon) \tilde{A}(s,a) \bigr)\Bigr] \\
      \log \pi(a) &= \log q(f_\theta(a)) + \sum\nolimits_j \log \vert  \frac{df_{\theta_j}(a)}{da} \vert \\
   f_{\theta_j}(a)_{d_j} &= a_{d_j} \\
   f_{\theta_j}(a)_{\backslash d_j} &= a_{\backslash d_j} \odot \exp(l\, \text{tanh}(s_{\theta_j}(a_{d_j}))) + t_{\theta_j}(a_{d_j})
\end{aligned}
\end{equation}
where $j$ represents the $j$th coupling layer. A pseudo-code of NFPO is presented in \Cref{app:pseudo_code}.

\section{Experiments}

In this section, we conduct experiments to answer several questions regarding NFPO's properties: 1) How do various factors influence NFPO's performance? 2) How is NFPO's performance compared to state-of-the-art (SOTA) PPO implementations? 3) Does NFPO learn multi-modal behaviors? 4) How is NFPO compared to other multi-modal policy learning methods?  5) Could NFPO's policy transfer to real-world robotic platforms? 
We conduct experiments in 2 widely used simulator: 1) Unitree RL Gym (URG) which is based on Nvidia IsaacGym \citep{makoviychukIsaacGymHigh2021b} and is the official platform for training locomotion policies on Unitree's robots (g1, h1 and go2); 2) Mujoco Playground (MJP) \citep{mujoco_playground_2025}  which uses MJX as the underlying parallel simulation.
\begin{figure}[h!]
   \centering
   \includegraphics[width=0.98\textwidth]{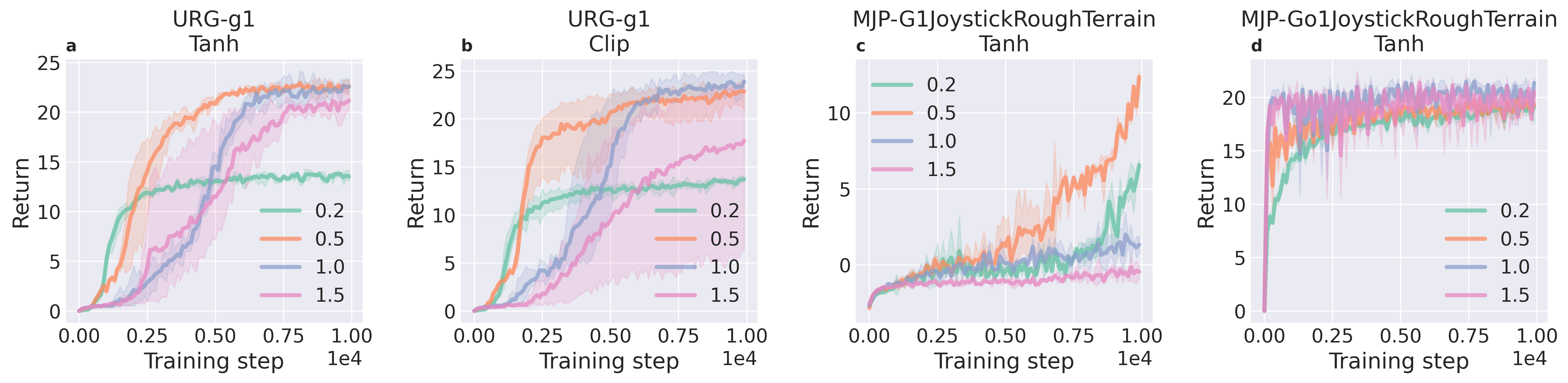}
   \caption{Studies in various factors. Experiments are in 3 seeds with 95\% confidence interval.}
   \label{fig:as1}
   
\end{figure}
\vspace{-10pt}
\subsection{Q1: How do various factors influence NFPO's performance?}
\label{sect:q1}
Our first experiment is  to explore how various factors influence the \texttt{NFPO}'s performance. As we have observed, the  `normalizing' technique is very important in stabilizing the training of NFs, we hence choose several related factors: 1) the hyperparameter $l$, 2) tanh vs clip normalization, 3) an entropy term that is used in PPO to prevent mode collapsing and 4) adding certain level of noise to actions during training then remove it in sampling phase, similar to \citet{zhaiNormalizingFlowsAre2025}. 

   \begin{wrapfigure}[27]{r}{0.5\textwidth}
      \vspace{-16pt}
   \centering
   \includegraphics[width=0.5\textwidth]{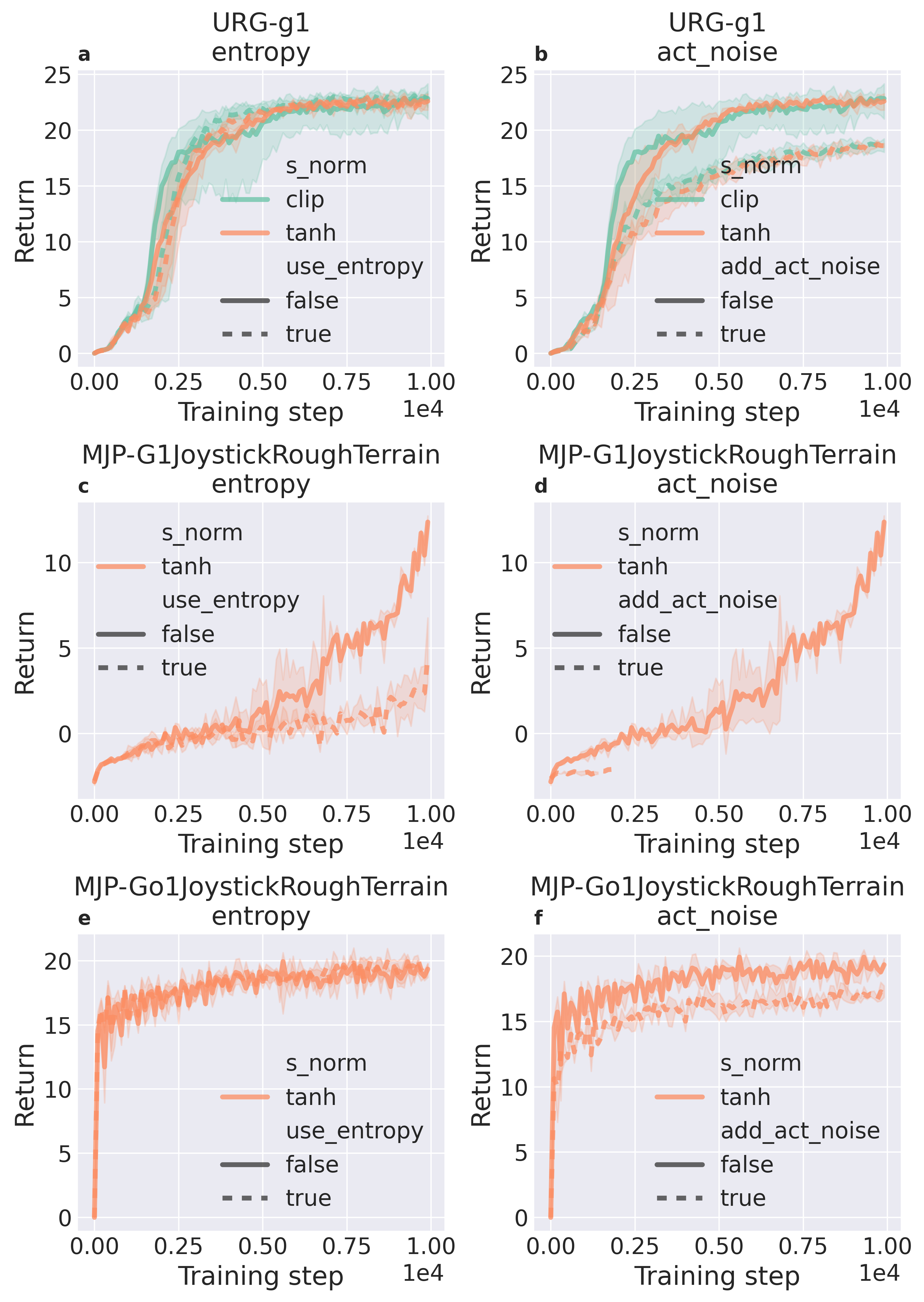}
   \vspace{-12pt}
   \caption{Studies in various factors. Experiments are in 3 seeds with 95\% confidence interval.}
   \label{fig:as2}
   \end{wrapfigure}
We firstly investigate the influence of `normalization' techniques between clip and tanh in \Cref{fig:as1} (subfigures \textbf{a} and \textbf{b}): Generally speaking, both clip and tanh could obtain strong performance given proper $l$ values. However, tanh tends to be more robust towards $l$ than clip hence is chosen in our implementation.

Then we check the influence of $l$ on tanh by looking at subfigures \textbf{a}, \textbf{c} and \textbf{d} in  \Cref{fig:as1}. In simpler environments like Go1JoystickRoughTerrain, various $l$ values could bring similar performance. But in challenging tasks like g1 and G1JoyStickRoughTerrain, proper value of $l$ plays a very important role in learning performance. From the results, we find $0.5$ is a good default value that fits various tasks.

Another important component in PPO is its entropy loss which could help exploration and prevent the methods from collapsing in local optimum. While \texttt{NFPO} is a multi-modal policy, we test if entropy term still helps its learning. By looking at results in \Cref{fig:as2} (subfigures \textbf{a}, \textbf{c} and \textbf{e}), we could find adding entropy loss in our \texttt{NFPO} doesn't bring significant performance difference and could even slow down the learning speed in G1JoyStickRoughTerrain.

Finally, as reported in \citet{zhaiNormalizingFlowsAre2025}, a critical technique to increase the image generation quality of NFs is to add Gaussian noise to training samples and remove this during generation, \textcolor{black}{which is also observed in \citet{ghugareNormalizingFlowsAre2025}}. While our online RL settings have offered massive amount of continuous samples that could potentially reduce the need to `dequantize' as in their  \textcolor{black}{dataset-based} settings, we still test if adding this training noise and remove it during action sampling could help us further improve the performance. The results are in \Cref{fig:as2} (subfigures \textbf{b}, \textbf{d} and \textbf{f}). Different from their settings,  we find adding action noise would only decrease the RL performance in our tasks. Especially in some challenging environment like G1JoyStickRoughTerrain, it triggers training instability and early stops training at roughly 25\% of total steps.

With above knowledge, we finally design our NFPO as a 4-layer RealNVP network with tanh transformations, remove entropy loss and use $l=0.5$. \textbf{We use this configurations in all following experiments, \textcolor{black}{replace Gaussian policy with it} and find this obtains competitive performance without further tuning \textcolor{black}{on other hyperparameters.}}.
\begin{figure}[h!]
   \centering 
   \includegraphics[width=0.99\linewidth]{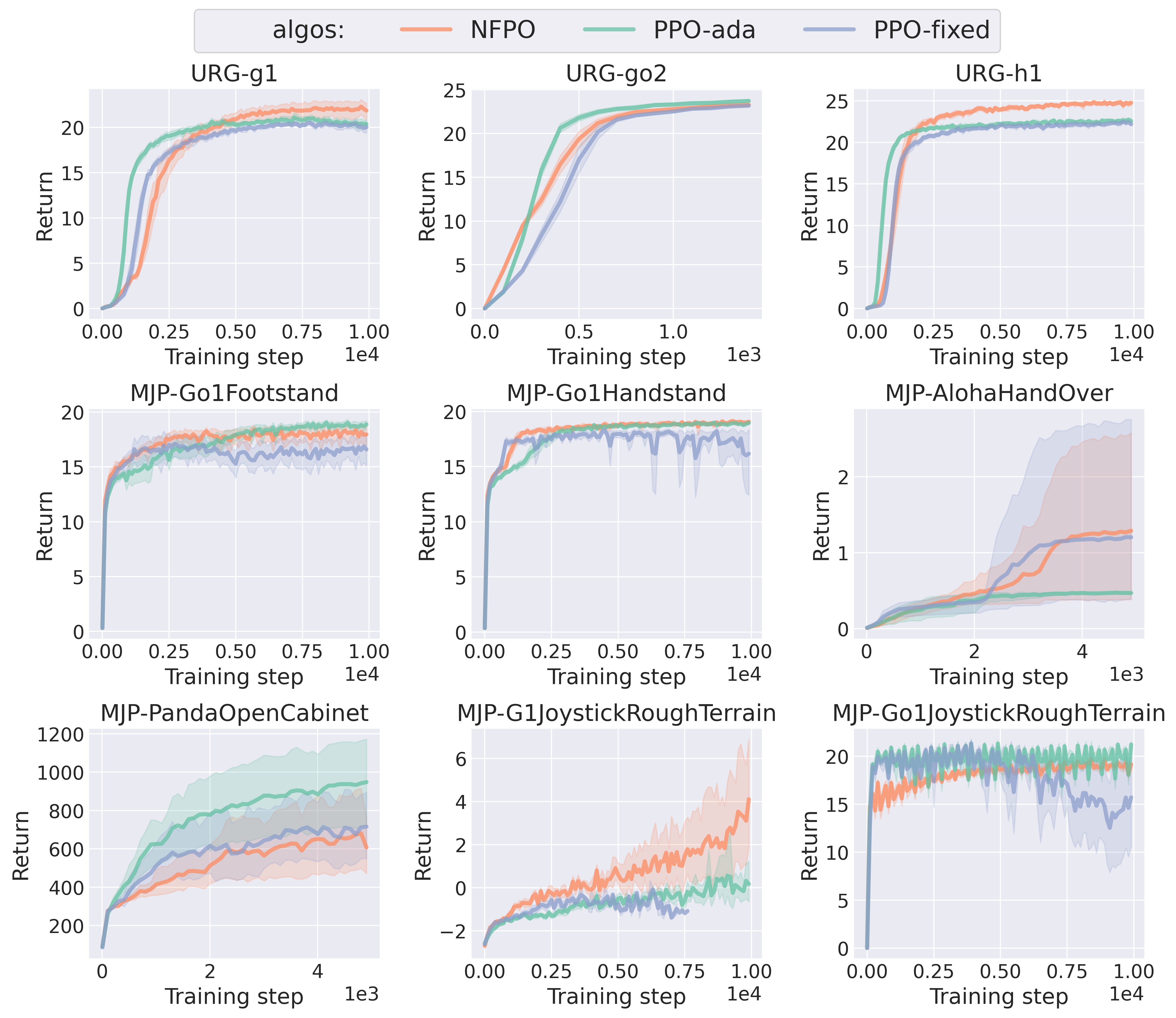}
   \caption{Learning curves of various methods on representative robotic tasks. The experiments are in 10 seeds. The errorbars are 95\% confidence interval.}
   \label{fig:fig1}
\end{figure}
\subsection{Q2: How is NFPO's performance compared to SOTA PPO implementations?}
In this section, we compare NFPO's performance against RSL-RL \citep{schwarke2025rslrl}'s PPO implementation, the latter is current SOTA method that widely used in real-world oriented policy learning in countless Robotics works.  We compare NFPO to 2 variants of PPO: 1) PPO-adaptive where the learning rate is dynamically scheduled for faster convergence and 2) PPO-fixed where the learning rate scheduler is disabled and is the default setting in Mujoco playground. The experiments are run in 10 seeds and shown in \Cref{fig:fig1}. For a fair comparison, we have aligned the actor network size of \texttt{PPO} and \texttt{NFPO}  and most other components (e.g., value network) and hyperparameters (e.g., learning rate) are shared. The result is in \Cref{fig:fig1}. We run 4096 parallel environments for unitree rl gym and 2048  for mujoco Playground.

From the result, we observe \texttt{NFPO} could achieve competitive or stronger performance compared to \texttt{PPO} baselines. Specifically, in high-dimensional control tasks like g1-joystick, h1-joystick and G1JoyStickRoughTerrain, our \texttt{NFPO} could achieve stronger convergence performance, thanks to its multi-modal modeling ability. On Go1JoystickRoughTerrain, \texttt{NFPO}'s performance fluctuates in less extent, reflecting it's better to adapt to complex terrains. For simpler tasks like go2-joystick, \texttt{NFPO} still obtains similar performance against \texttt{PPO}. Finally, in manipulation tasks like AlohaHandOver and PandaOpenCabinet where exploration plays a more important part, the three  tested methods exhibit complicated result with no one to be the best, indicating careful tuning is needed in these tasks. As for computation efficiency, on Unitree gym's g1 environment, we observed a roughly 19\% increase of wall-clock time in \texttt{NFPO} with details in \Cref{app:runtime}. For tasks in mujoco playground, we provide a video of \texttt{NFPO}'s policies in simulation in supplemental materials.

\subsection{Q3: Does NFPO learn multi-modal behaviors?}
In this section, we test if \texttt{NFPO} could learn multi-modal behaviors given proper reward functions. Following works in FPO \citep{mcallisterFlowMatchingPolicy2025}, we use the gridworld example to demonstrate the difference in generated behaviors between \texttt{NFPO} and \texttt{PPO}.
In gridworld environment, the agent is spawned in the grey cells and take continuous actions to move around. If it hits the green region, a positive reward would be granted otherwise the reward is 0.

   \begin{wrapfigure}[18]{r}{0.44\textwidth}
      \vspace{-16pt}
   \centering
   \includegraphics[width=0.44\textwidth]{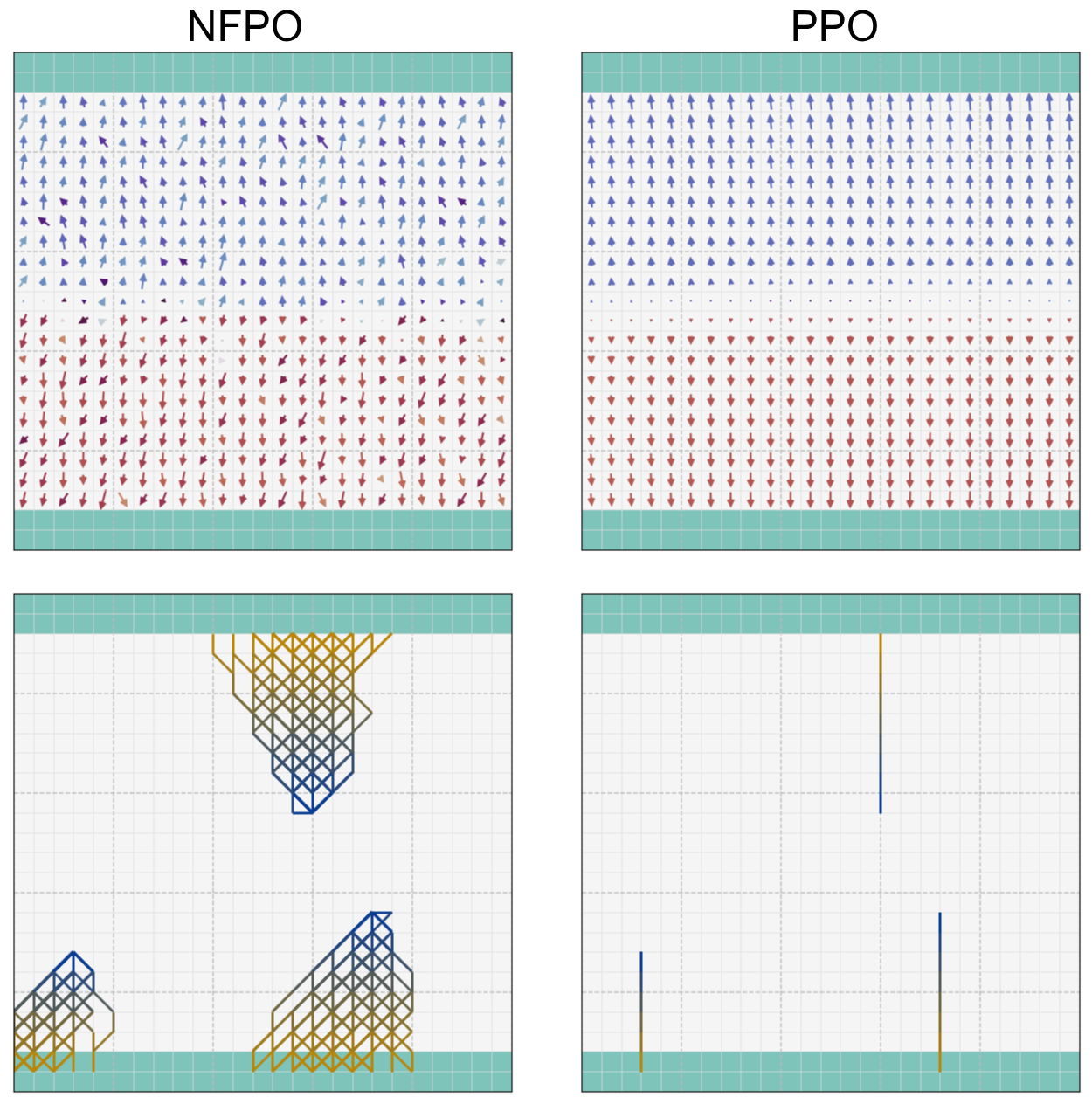}
   \vspace{-15pt}
   \caption{Difference in generated trajectory of \texttt{NFPO} and \texttt{PPO} in gridworld. }
   \label{Fig:multi_modal}
   \end{wrapfigure}
The result is shown in \Cref{Fig:multi_modal}. The upper 2 figures depict the actions each policy would take in certain position and the lower 2 figures depict the generated 100 trajectories from the same starting point of each method. In this sparse reward setting, we could find \texttt{NFPO} tends to generate diverse actions that lead to various trajectories towards green region while \texttt{PPO} only takes the shortest way and doesn't exhibit multi-modal behaviors.

\begin{wrapfigure}[8]{r}{0.5\textwidth}
   \vspace{-18pt}
\centering
\includegraphics[width=0.24\textwidth]{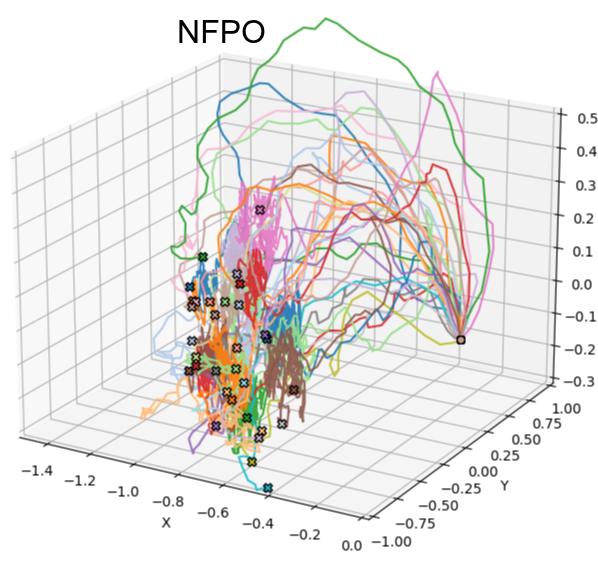}
\includegraphics[width=0.24\textwidth]{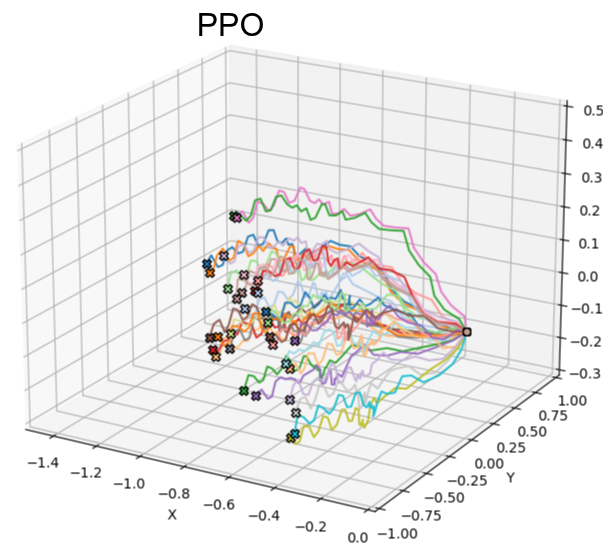}\\
\vspace{-5pt}
\caption{Difference in generated trajectories of \texttt{NFPO} and \texttt{PPO} in UR10 Reaching.}
\label{Fig:multi_modal2}
\end{wrapfigure}
Besides, we also train \texttt{NFPO} and \texttt{PPO} in Isaac-Reach-UR10-v0 environment in IsaacLab \citep{mittal2023orbit}. Specifically, we design a sparse-reward variant of UR10 reaching where the arm would gain a positive reward when reaching to the given target position.
After training both \texttt{PPO} and \texttt{NFPO} to convergence, we fix the initial state  and target, then run both \texttt{PPO} and \texttt{NFPO} for 100 rollouts, and plot their trajectories in \Cref{Fig:multi_modal2} where `o' marks the initial position (in xyz world frame) of end effector and `x' marks the end position. We also attach the videos captured in IsaacLab's simulation environment to better visualize the behavior difference between \texttt{NFPO} and \texttt{PPO} in supplemental materials.

From the result, we could find \texttt{NFPO} generates diverse trajectories while \texttt{PPO} only runs in rather fixed trajectories. 

\subsection{Q4: How is NFPO compared to other multi-modal policy learning methods?}
In this section, we try to compare \texttt{NFPO} against several related works that also employ multi-modal policy parameterizations in onlint RL setting. However, as also illustrated in \Cref{tbl:algo_compare}, not all methods hold equal fitness and potentials to be used in robotic policy learning. Specifically, there are 2 lines of works that deserve a discussion and comparison:

\begin{enumerate}[leftmargin=*]
   \item On-policy diffusion-model-based methods like GenPO \citep{dingGenPOGenerativeDiffusion2025} and FPO \citep{mcallisterFlowMatchingPolicy2025}. The difference between \texttt{NFPO} to these methods mainly locate in the difference between diffusion-based method and normalizing flows: In NFs, we have closed-form, easily-calculated log probability that could be directly chained to PPO's training objective. While for diffusion models, various approximation methods need to be used and incur memory and computation overhead. As GenPO doesn't release their source code and is hungry in memory and computation resources, we compare against FPO in this genre of methods. 
   \item Off-policy normalizing-flows-based methods like Meow \citep{chaoMaximumEntropyReinforcement2024} and \citet{ghugareNormalizingFlowsAre2025}. The difference between \texttt{NFPO} to these methods mainly locate in the training paradigm of PPO to off-policy methods: Off-policy methods usually have a larger replay buffer and better sample-efficiency but fall short of computation and memory efficiency as also studied in \citet{seoFastTD3SimpleFast2025}. In this line of work, we compare against Meow as \citet{ghugareNormalizingFlowsAre2025} doesn't directly study conventional online RL settings.
\end{enumerate}

As for other off-policy diffusion-based methods like MaxEntDP \citep{dongMaximumEntropyReinforcement2025}, QVPO \citep{dingDiffusionbasedReinforcementLearning2024} and QSM \citep{psenkaLearningDiffusionModel2024}, both their off-policy setting and approximation of log probability makes it non-trivial to integrate into normal training pipelines like IsaacGym and Mujoco Playground hence are omitted here.

\begin{figure}[h!]
\centering
\includegraphics[width=0.98\textwidth]{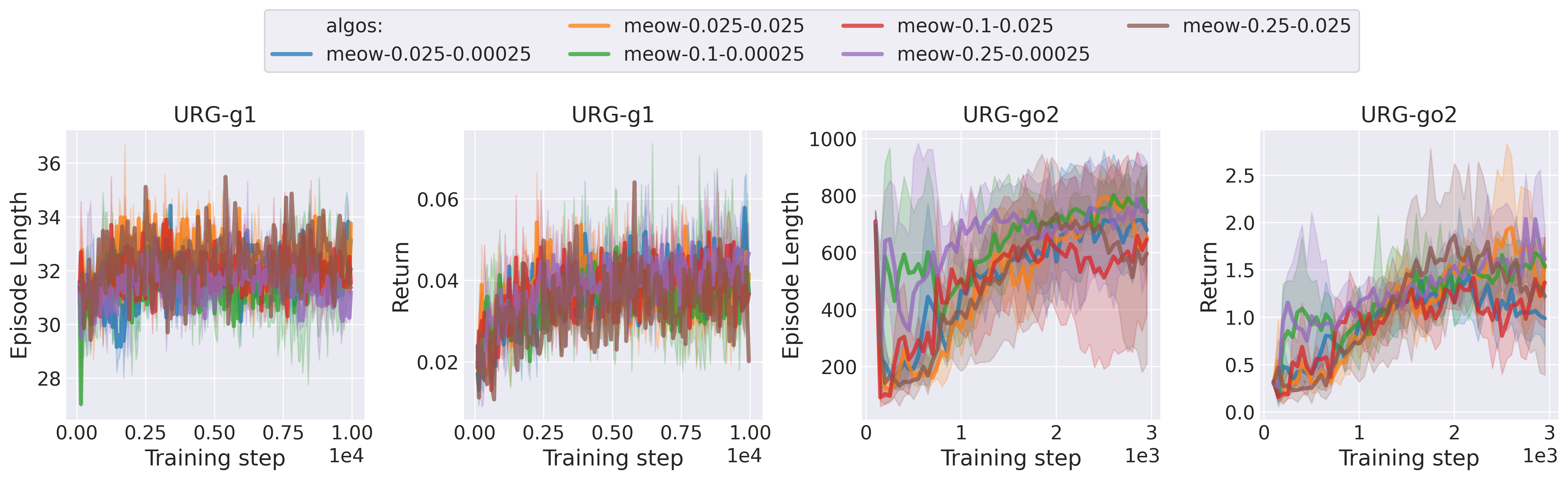}
\caption{Learning curves of \texttt{Meow} and \texttt{NFPO} in 2 environments of Unitree RL Gym. \textcolor{black}{Legend is meow-\texttt{alpha}-\texttt{polyak} where \texttt{alpha} and \texttt{polyak} are 2 tuned hyperparameters.}}
\label{Fig:baseline}
\end{figure}

In \Cref{Fig:baseline}, we present the training curves of Meow in Unitree RL gym's g1 and go2 environment. In \texttt{Meow}, it has a novel design where no explicit policy network exists. Rather, it uses value function $Q$ to parameterize the policy but this conflicts with the common asymmetric setting in robotic tasks where the value network and actor network have different observation dimensions and we have to feed actor observations (without privileged information) to its $Q$ and $V$ functions. Besides, we adjust its off-policy replay buffer's dimension to $(N_e, N_r, ...)$ where the first dimension is the number of environments (4096 or 2048), the second dimension is its original buffer length (1.5e4). Following \texttt{FastTD3}, we also increased the batch size per each update while all other settings are either from their original implementation or the same to \texttt{PPO} and \texttt{NFPO} (\textcolor{black}{details are in \Cref{app:tune_meow}}). Under this setting, it costs roughly 50GB GPU memory for training on 4096 unitree gym's g1 environment.

\begin{wraptable}[13]{r}{7.8cm}
   \centering
   \vspace{-14pt}
      \caption{\texttt{NFPO} and \texttt{FPO}'s accomplished tasks in Mujoco playground.}
      \label{tbl:mjp_rlt}
   \begin{threeparttable}
      \newcolumntype{C}[1]{>{\centering\arraybackslash}m{#1}}

        \begin{NiceTabular}{|C{3.9cm}|C{1.0cm}|C{1.0cm}|}[hvlines]
            \CodeBefore
                \rowcolor{lightgray}{1-1}
            \Body
            Task & FPO& NFPO\\

            AlohaHandOver & \cmark & \xmark  \\
            G1JoystickRoughTerrain & \xmark & \cmark  \\
            Go1Footstand & \xmark & \cmark  \\
            Go1Handstand & \cmark & \cmark  \\
            Go1JoystickRoughTerrain & \xmark & \xmark  \\
            PandaOpenCabinet & \xmark & \cmark  \\
            PandaPickCube & \xmark & \cmark  \\
            Total & 2 &  5  \\
            
        \end{NiceTabular}
    \end{threeparttable}
\end{wraptable}
From the result, unfortunately, \texttt{Meow} doesn't  exhibit meaningful performance in g1 and go2 environments. In g1, both the episode length and episodic return stay in low values while in go2, episode length achieves high values in some phase, indicating \texttt{Meow}'s maximum-entropy learning framework could incentivize the exploration in some extent. However, the episodic return stays in low values and do not indicate a successful learning.

For \texttt{FPO}, we use their original codebase, and run the \texttt{FPO}'s training with their default hyperparameters on 7 mujoco playground's tasks similar to what we used in \Cref{fig:fig1}. Then we visualize the learned policies in official mujoco playground's codebase to see if they could successfully perform the tasks and show results in \Cref{tbl:mjp_rlt}.  For \texttt{FPO}, 2 out of 7 policies could show meaningful behaviors in simulation. And \texttt{NFPO} could perform 5 out of 7.  Videos of both \texttt{NFPO} and \texttt{FPO} could be found in supplemental materials.

Integrating modern multi-modal policy into common robotic learning tasks is not a straightforward work, as the robotic tasks feature learning stability and memory and computation efficiency more than traditional RL tasks. And the purpose of this experiment is to showcase the learning stability and robustness of \texttt{NFPO} for which we use \textit{a same set of configuration} across multiple simulation environments and tasks. It's also important to note this experiment doesn't mean \texttt{FPO} and \texttt{Meow} \textit{won't work in robotic policy learning}. Actually, with proper hyperparameter tuning and training adjustment, it's likely for these methods to obtain improved performance, as in \citet{seoFastTD3SimpleFast2025}. %

\subsection{Q5: Could NFPO's policy transfer to real-world robotic platforms?}
\begin{figure}[t]
\centering
\includegraphics[width=0.16\textwidth,height=0.25\textwidth]{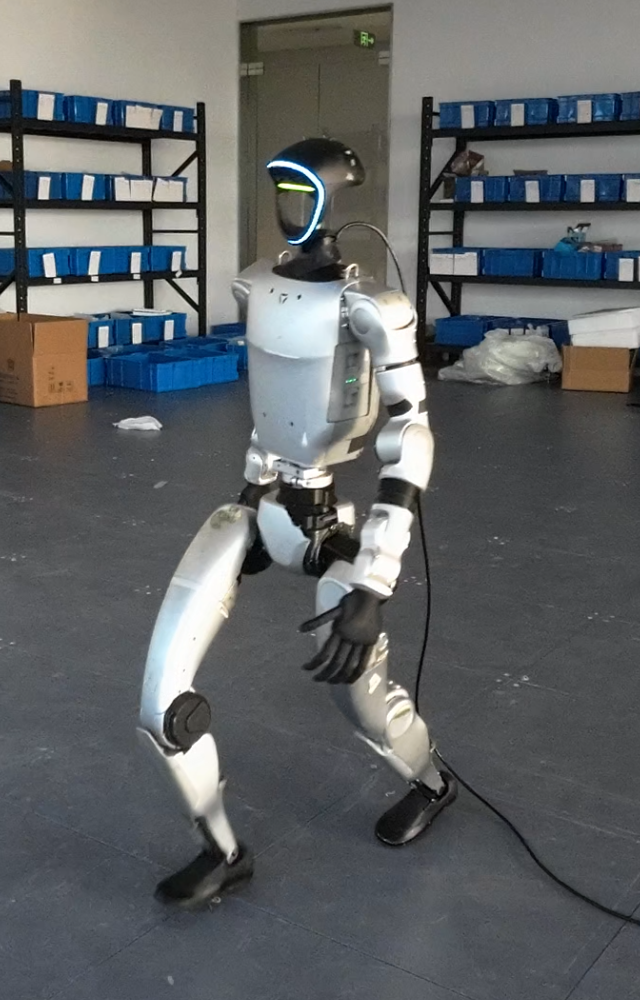}
\includegraphics[width=0.16\textwidth,height=0.25\textwidth]{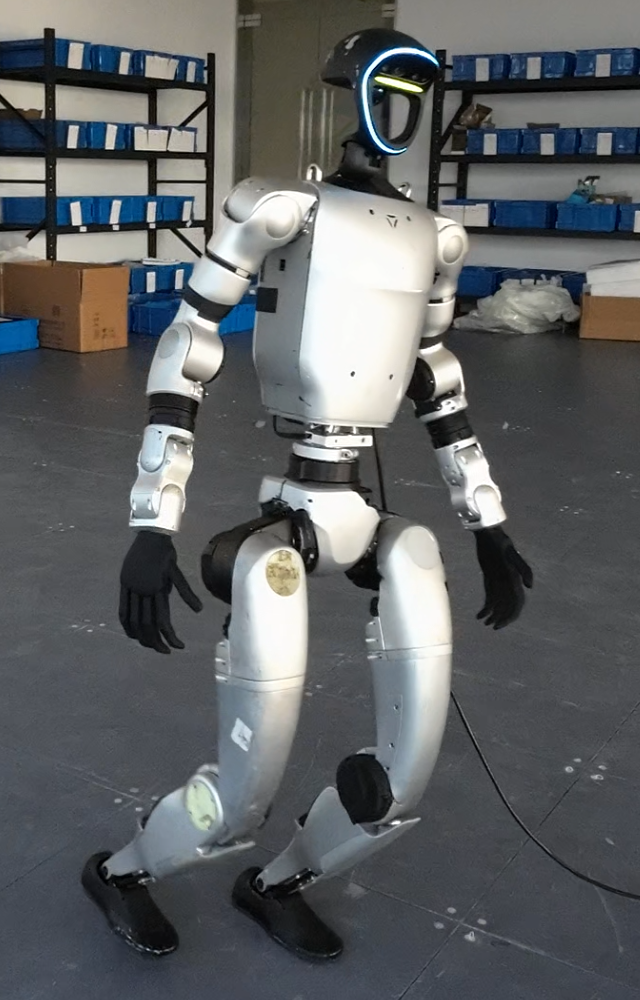}
\includegraphics[width=0.16\textwidth,height=0.25\textwidth]{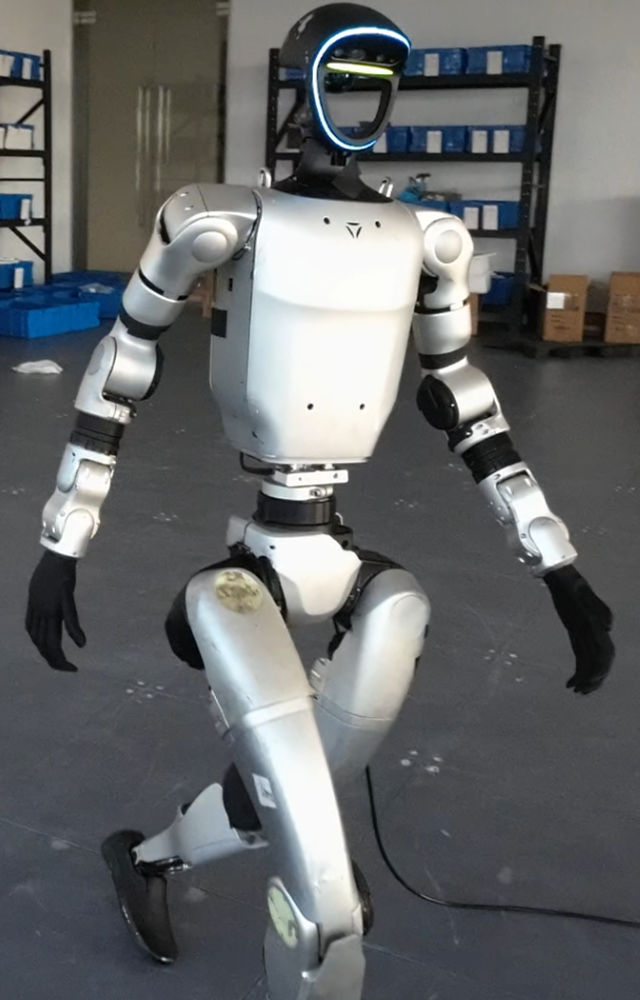}
\includegraphics[width=0.16\textwidth,height=0.25\textwidth]{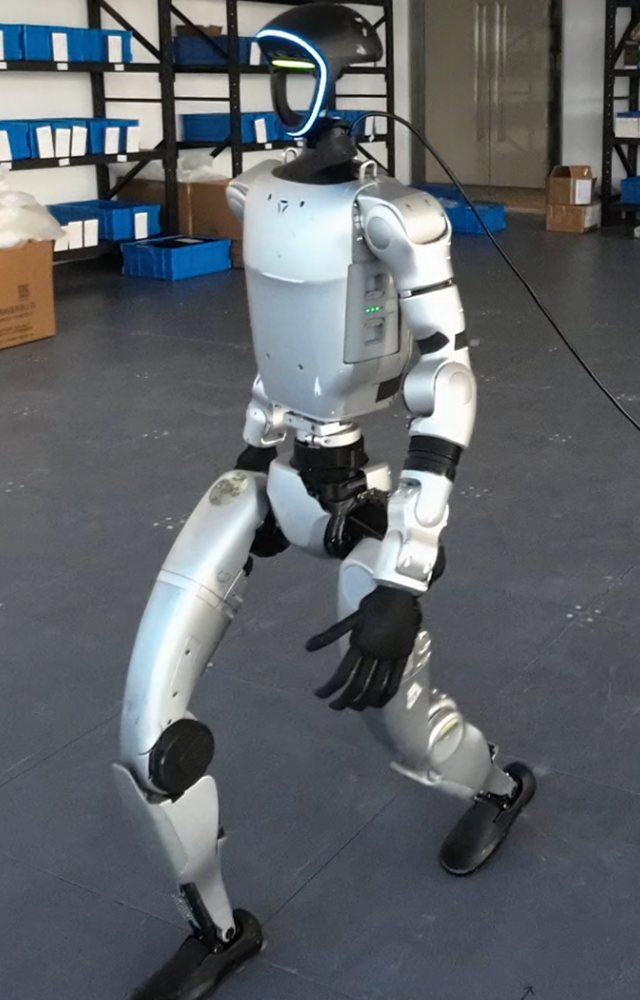}
\includegraphics[width=0.16\textwidth,height=0.25\textwidth]{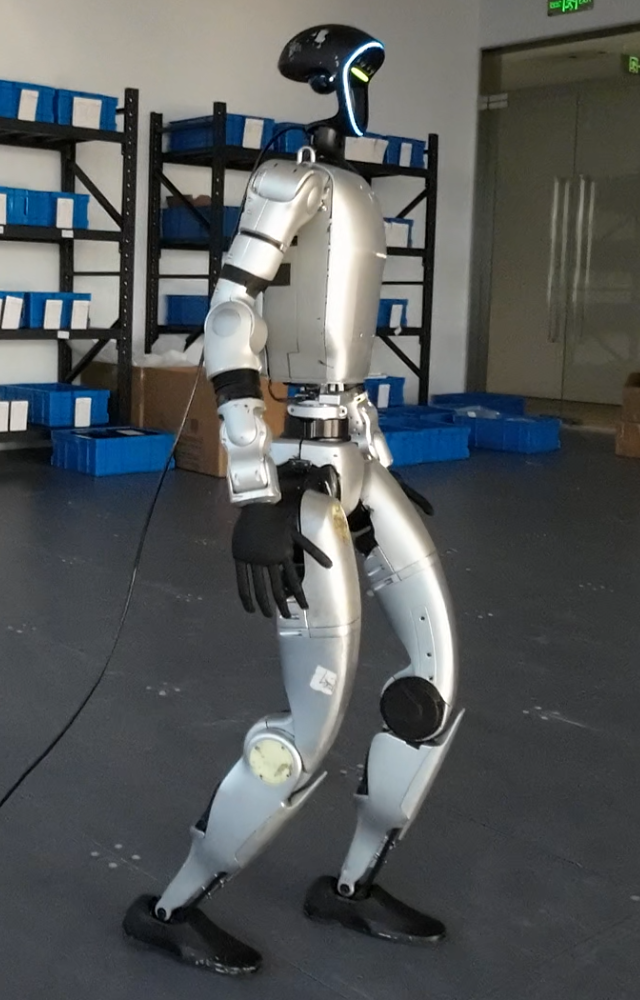}
\includegraphics[width=0.16\textwidth,height=0.25\textwidth]{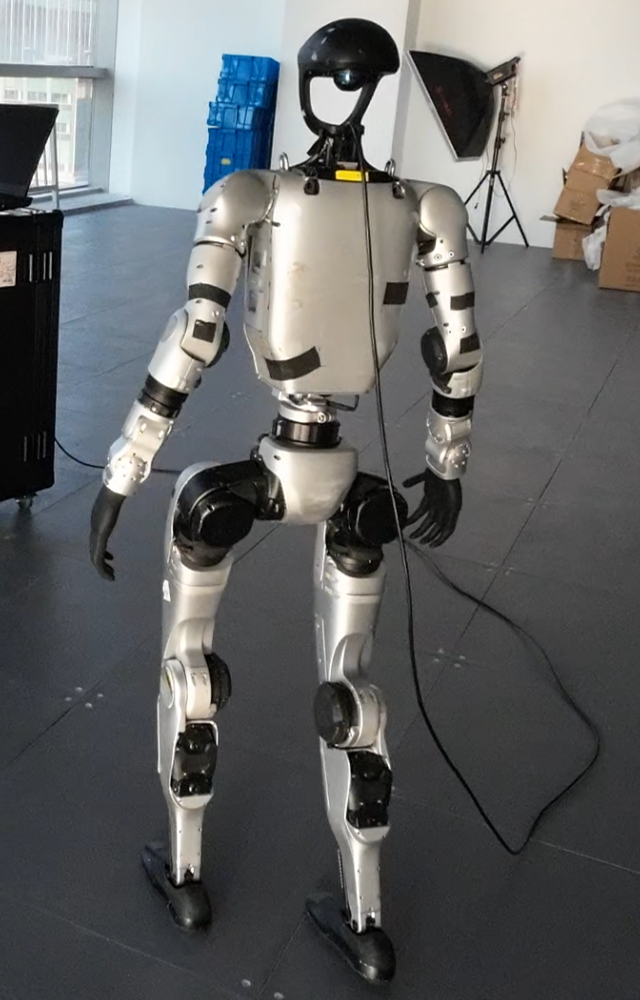}\\

\includegraphics[width=0.16\textwidth,height=0.25\textwidth]{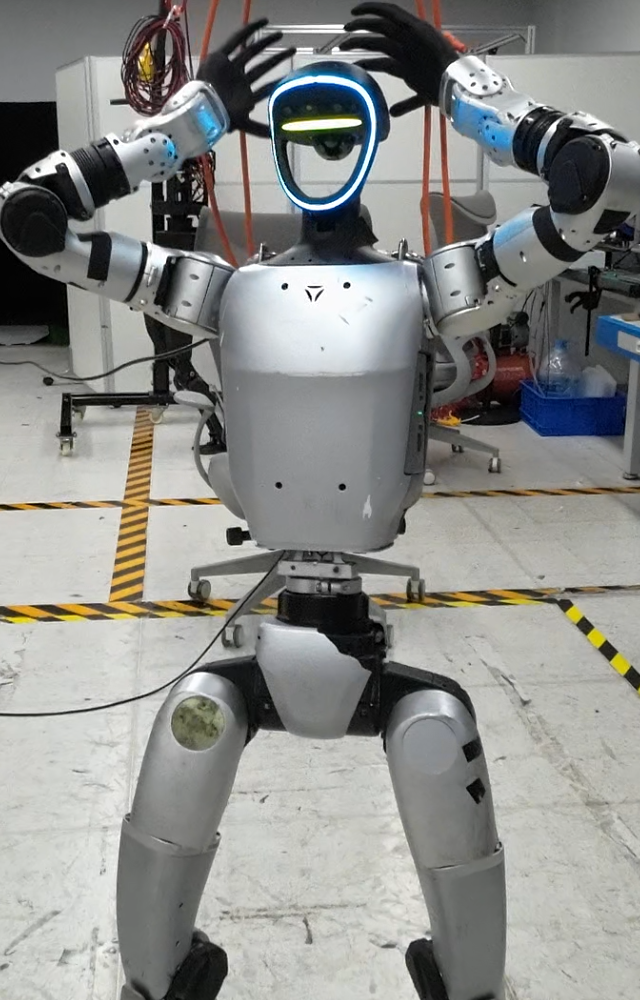}
\includegraphics[width=0.16\textwidth,height=0.25\textwidth]{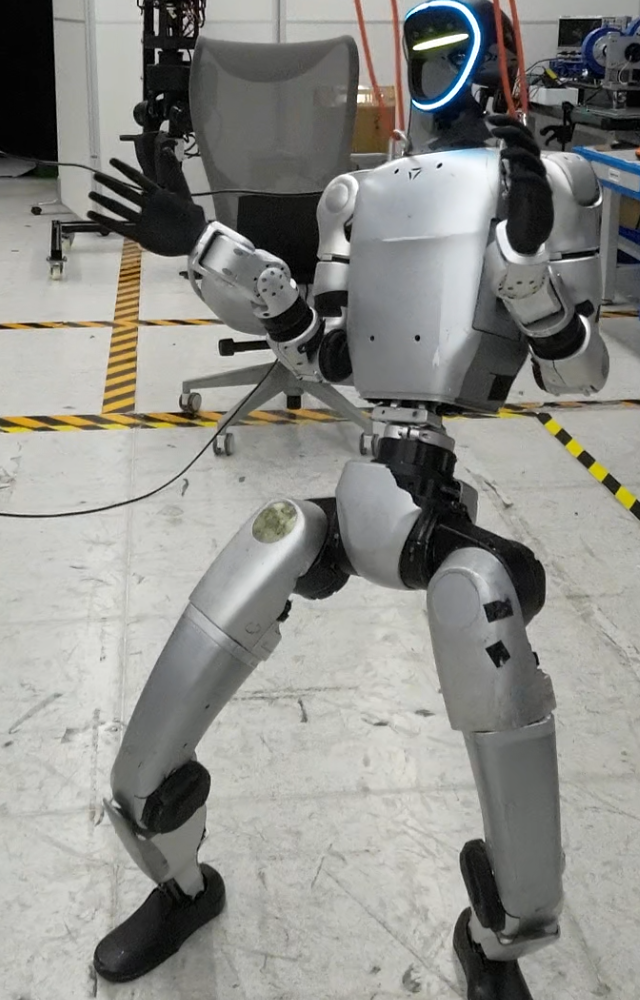}
\includegraphics[width=0.16\textwidth,height=0.25\textwidth]{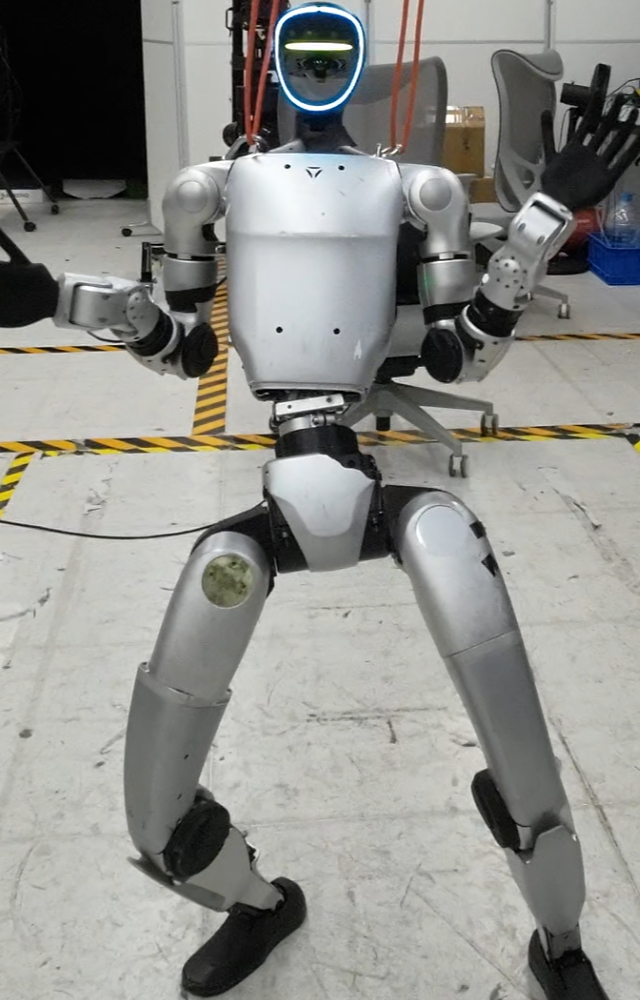}
\includegraphics[width=0.16\textwidth,height=0.25\textwidth]{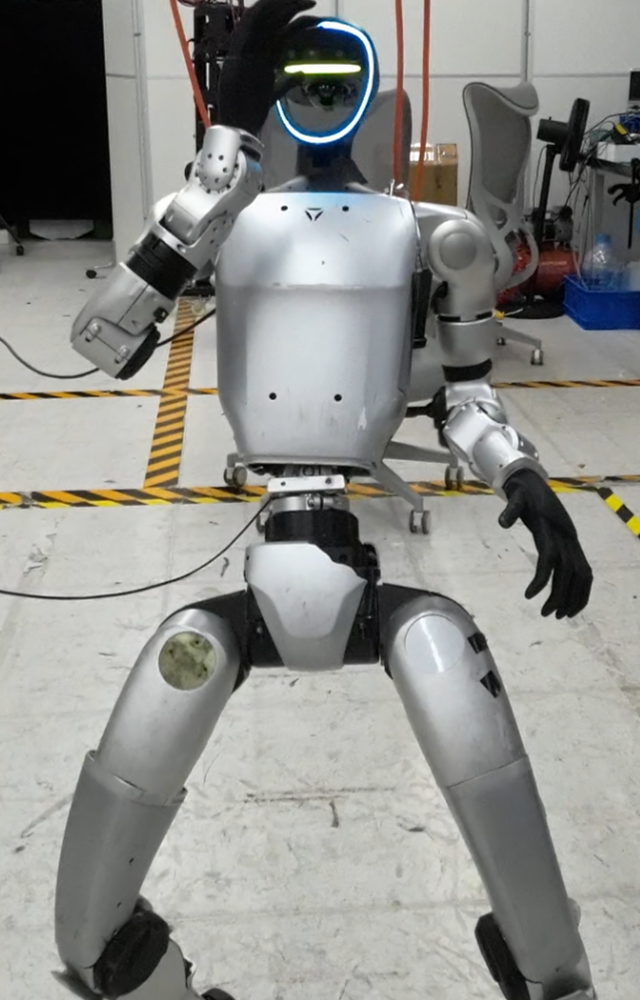}
\includegraphics[width=0.16\textwidth,height=0.25\textwidth]{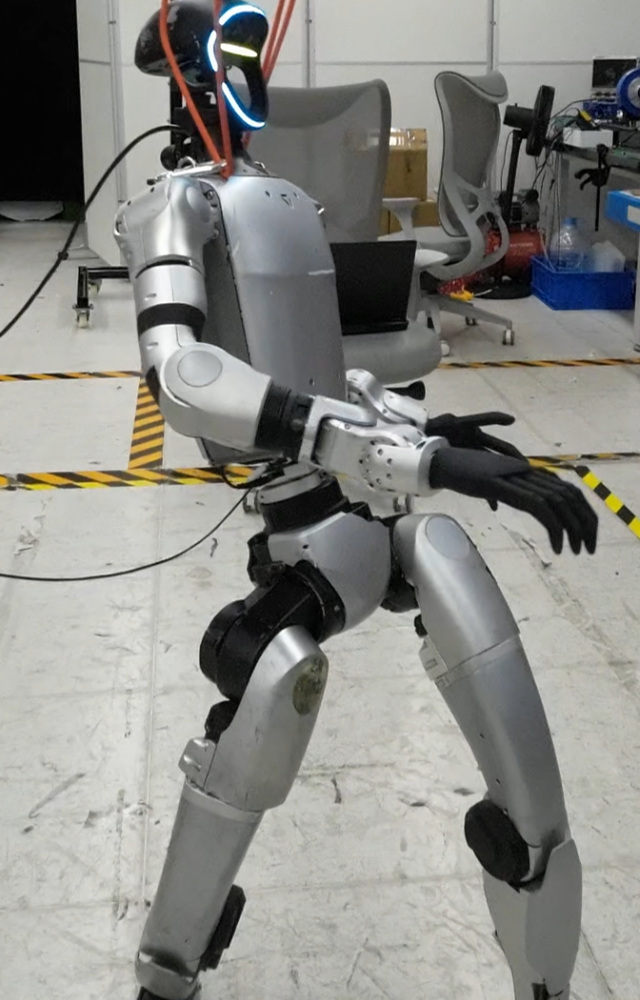}
\includegraphics[width=0.16\textwidth,height=0.25\textwidth]{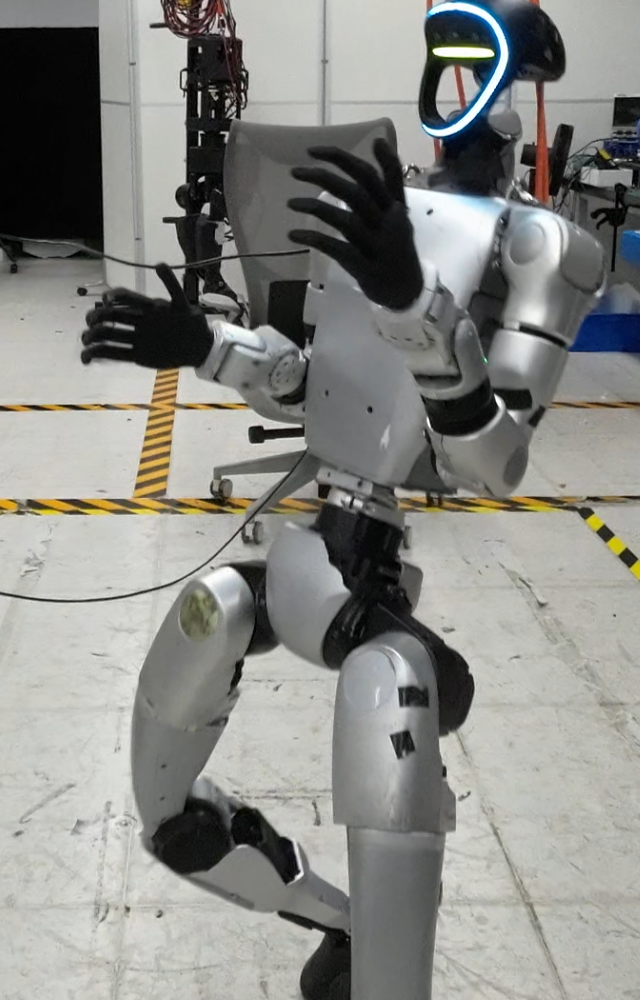}
\caption{Snapshoots of real world deployment video.}
\label{Fig:real_world1}
\end{figure}

In this section, we transfer the policies trained by \texttt{NFPO} onto real-world robotic platform. In details, we choose Unitree RL Lab \citep{unitree_rl_lab_2025} for training a joystick locomotion policy of and BeyondMimic \citep{liaoBeyondMimicMotionTracking2025} to train a dancing policy via motion-tracking.  \textcolor{black}{The robot is Unitree's g1 and we use the common hierarchical structure where a high-level RL policy (trained by \texttt{NFPO}) outputs action target then a low-level PD controller outputs joint torque accordingly. The high-level policy runs in 50Hz and the low level PD controller runs in 200Hz.  The video} of real world deployment is in supplemental materials and a series of the video clips is provided in \Cref{Fig:real_world1}. Similar to \citet{chaoMaximumEntropyReinforcement2024}, we have found deterministic version of \texttt{NFPO} could generate stable and smooth real world motions than its stochastic counterpart (A discussion is provided in \Cref{app:deploy}).
From the results, we could find the policy trained in \texttt{NFPO}  could successfully perform stable and agile \textcolor{black}{29-DoF whole-body control on unitree's G1 robot}. 

\section{Conclusion}
In this work, aiming to unlock DRL-based multi-modal robotic policy learning, we explore the integration of Normalizing Flows to online Policy Optimization framework and have successfully built NFPO which is a stable method and could transfer to real-world robotic tasks. 

In future works, we aim to further explore the benefits of  NFPO to robotic policy learning, for eaxmple how to take advantages of its efficient calculation of log probability for interpretation and optimization of neural network based robotic policy.

\newpage

\bibliography{iclr2026_conference}
\bibliographystyle{iclr2026_conference}

\newpage
\appendix
\section{Appendix}
\setcounter{figure}{0}
\setcounter{table}{0}
\setcounter{equation}{0}
\renewcommand{\thetable}{S\arabic{table}}
\renewcommand{\theequation}{S\arabic{equation}}
\renewcommand{\thefigure}{S\arabic{figure}}
\subsection{Determination of values in \Cref{tbl:algo_compare}}
\label{app:table_details}
In this section, we explain how we choose the values for each algorithm presented in \Cref{tbl:algo_compare}. We focus on explaining those that may cause confusion and omit the easy ones.

\begin{enumerate}
   \item FastTD3 \citep{seoFastTD3SimpleFast2025}. In their paper, a critical modifications to the TD3 algorithm is to increase its batch size to a rather large value (roughly 32k). Combined with large off-policy replay buffer and massive environment numbers, the memory efficiency is sacrificed.
   \item MaxEntDP \citep{dongMaximumEntropyReinforcement2025}. In their paper, to compute the log probability of diffusion policies (Eq. 21 in the original paper) and estimation of training target (Eq. 17 in the original paper), sampling-based approximation is employed. In high-dimensional space, the sampling number $N$ and $K$ is expected to be large and they use $N=100$ and $K=500$ as noted in their paper. In our setting, combined with large number of parallel environments (typical value 2048, 4096), this results in high consumption of memory.

   \item GenPO \citep{dingGenPOGenerativeDiffusion2025}. GenPO tries to compute the full Jacobian determinant without simplification used in Normalizing Flow and this brings much computation and memory burden. As noted in Conclusion and Limitation of their paper: `\textit{Despite its excellent performance, GenPO faces the problem of relatively high computational and memory overhead to be resolved in the future.}'. And they used computing clusters with high memory and computation capacity as noted in their Appendix B.1.
\end{enumerate}

The references: 1) FastTD3: \citep{seoFastTD3SimpleFast2025}; 2) Meow: \citep{chaoMaximumEntropyReinforcement2024}  3) MaxEntDP: \citep{dongMaximumEntropyReinforcement2025};  4) GenPO: \citep{dingGenPOGenerativeDiffusion2025}; 5) FPO: \citep{mcallisterFlowMatchingPolicy2025}

\subsection{\textcolor{black}{Analysis on the stability of tanh vs clip}}
\label{app:theory_of_tanh_clip}
In this section, we provide a theoretical analysis on why tanh may perform better than clip in mitigating the training instability.

Firstly, let's recall we are maximizing %

$J(\theta) \approx  \mathbb{E} [r(\theta)\tilde{A}(s,a)] \approx \mathbb{E} [ \pi_\theta(a; s)] \approx  \mathbb{E}\biggl[\exp\Bigl(g\bigl(s_\theta(x)\bigr)\Bigr)\biggr] $ where $g$ is  one of \textit{identity function} ($g(x)= x$), \textit{hard clip} ($g(x; l) = \text{clip}(x, l)$) and \textit{tanh} ($g(x; l) = l \text{tanh}(x)$) ($l$ would be omitted for brevity in below).

If we analyze per-sample gradient of $J$ against $\theta$, we have:
\begin{align}
  G(x) = \nabla_\theta [\exp(g(s_\theta(x)))]  = \exp(g(s)) g^\prime(s) \nabla_\theta s
\end{align}

and the per-sample gradient scale is determined by multiplication  of 3 factors:
\begin{enumerate}
   \item $\exp(g(s))$: an exponential scale over $g$
   \item $g^\prime(s)$: how much the signal of outer exponential passes through
   \item $\nabla_\theta(s)$: the neural network's sensitivity towards its parameters and are not influenced by the choice of  $g$.
\end{enumerate}

\textbf{Identity Function}. For $g(x) = x$, we have:
$G(x) = \exp(s)\nabla(s)$ where $\exp(s)$ is unbounded and large $s$ could cause exploding gradients, same as what we observed in \Cref{Fig:motiv}.

\textbf{Hard Clip}. For $g(x; l) = \text{clip}(x, l)$, we have:
$G(x) = \exp(\text{clip}(s)) g^\prime(s) \nabla(s)$. While it helps bound the exponential value, there are 3 minor issues related to it: 1) When $s$ is inside the range, the training signal is  $\exp(s)$ which amplifies the gradient for larger $s$, creating a self-reinforcing push toward even larger values of $s$; 2) When s is outside clipped range, the gradient is 0, i.e. no training signal is received from this sample. Hence the optimizer loses ability to move those samples back;  3) Together,  many samples are driven toward the boundary $\vert s \vert = l$,  once they cross they become gradient-dead under hard clipping, which can stall or bias training.

\textbf{Tanh}. We have $G(x) = \exp(\text{tanh}(s))(1-\text{tanh}^2 s)\nabla s$. Besides bounded first term $\exp(\text{tanh}(s))$, another good property is the second term $(1-\text{tanh}^2 s)$ is also smoothly decreasing as $s$ increases. This helps optimizer to softly attenuate extreme values instead of zeroing them, so it avoids both exploding gradients and the “dead-zone” stalls caused by hard clipping.

\subsection{\textcolor{black}{Additional experiments on Mujoco}}
\label{app:mujoco_exp}
In this section, we compare \texttt{NFPO} to various classic online RL methods on Mujoco \citep{mujoco} environments, we use \texttt{CleanRL} \citep{huang2022cleanrl} as the testbed. For \texttt{SAC}, \texttt{TD3} and \texttt{PPO}, we use the original  implementation and hyperparameters in \texttt{CleanRL}'s codebase. For \texttt{NFPO}, we just replace the policy parameterization from \texttt{PPO} and others remain the same. Note as \texttt{SAC}, \texttt{TD3} are off-policy methods, \textit{they generally obtain better sample efficiency} compared to \texttt{PPO} and \texttt{NFPO}.

\begin{figure}[htbp]
   \centering 
   \includegraphics[width=0.99\linewidth]{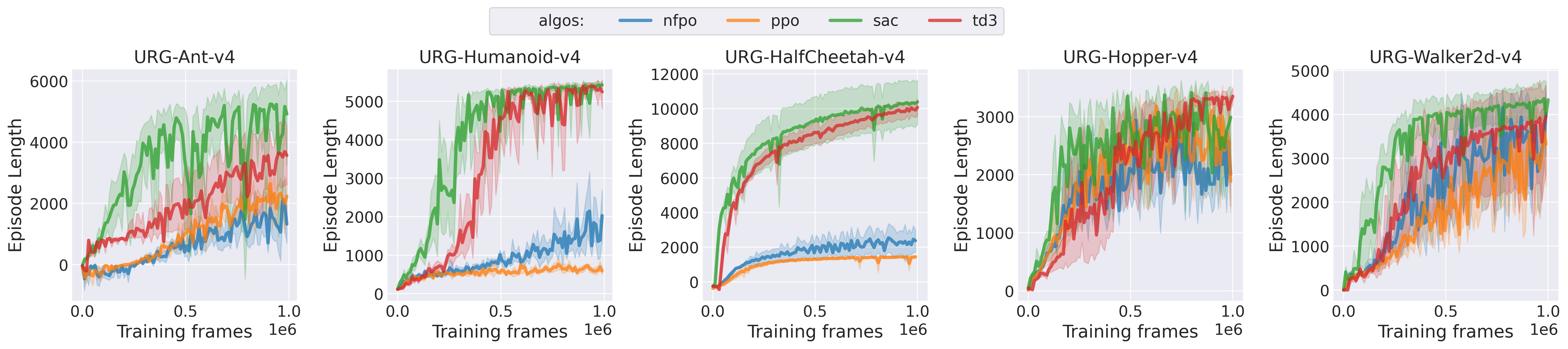}
   \caption{Performance of various online RL methods on Mujoco environments (5 seeds).}
   \label{fig:mujoco}
\end{figure}

From the results, we could find \texttt{NFPO} could obtain similar performance compared to \texttt{PPO} on various simpler environments like Ant, HalfCheetah, Hopper and Walker2d. For complex environments like Humanoid, it achieves better performance compared to \texttt{PPO}, similar to what we've found in the main experiments.

\subsection{\textcolor{black}{Additional ablation experiments of \texttt{NFPO}}}
\begin{figure}[htbp]
   \centering 
   \includegraphics[width=0.445\linewidth]{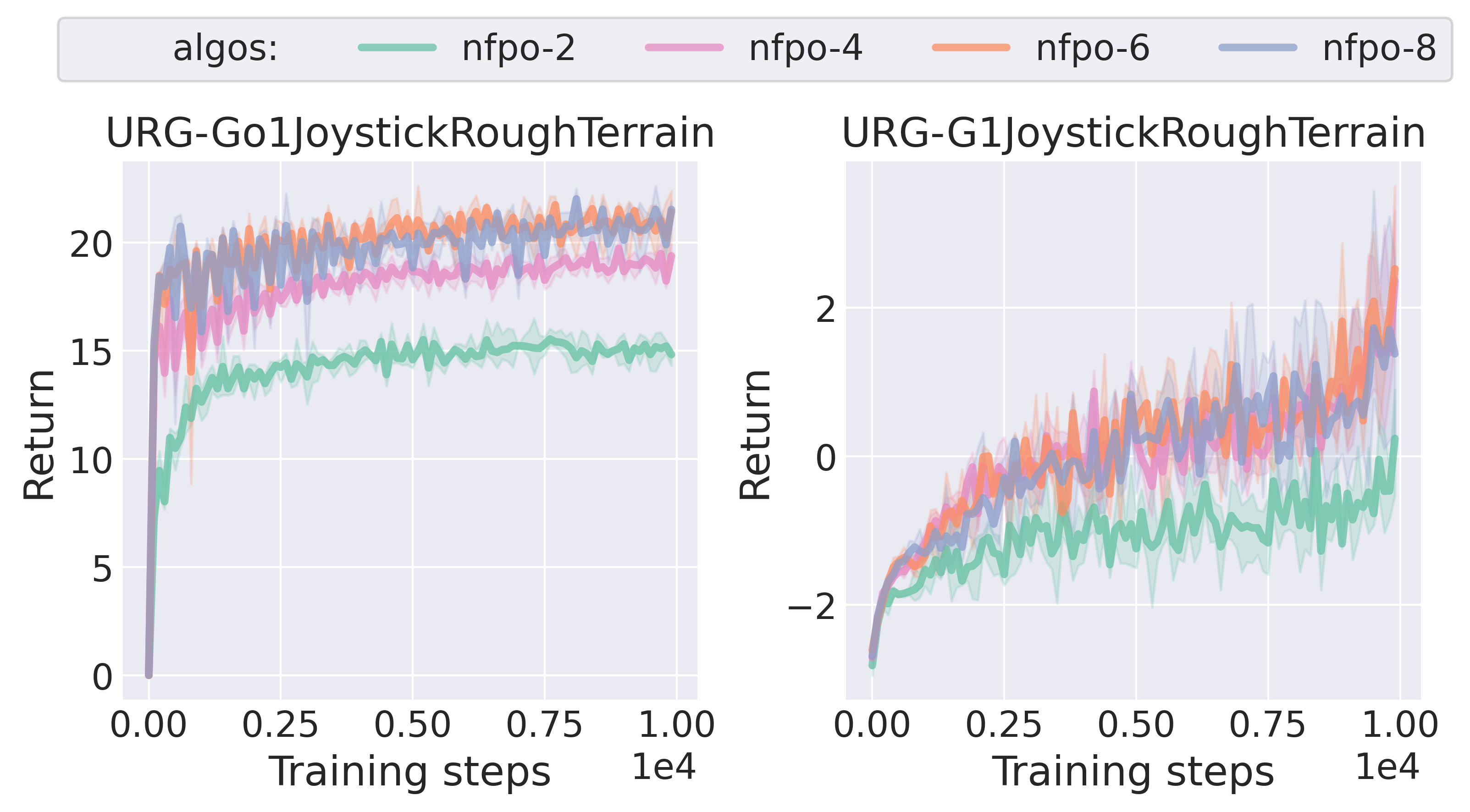}
   \includegraphics[width=0.48\linewidth]{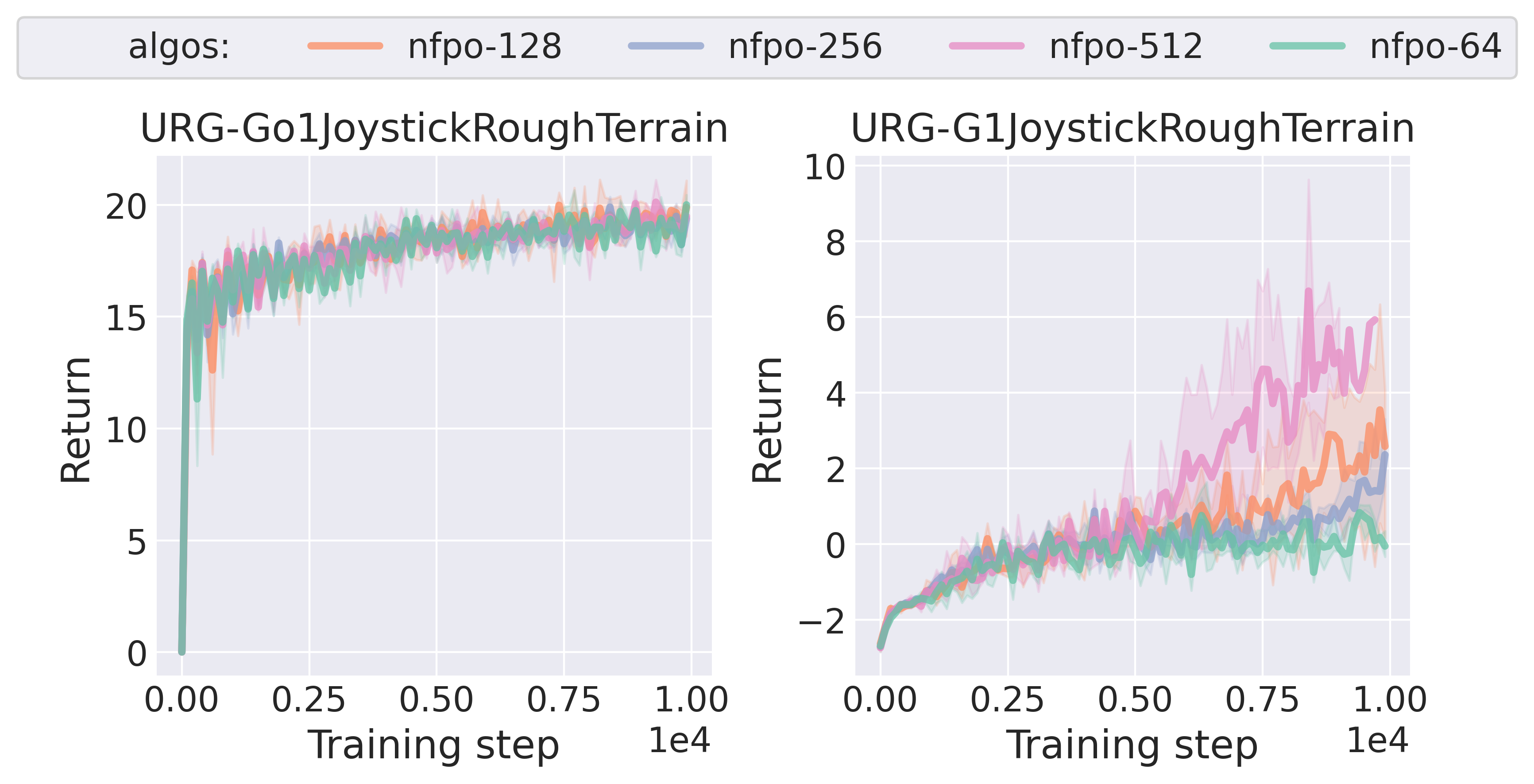}
   \caption{Performance of \texttt{NFPO} over various number of layers and hidden dimensions (5 seeds).}
   \label{fig:as_more}
\end{figure}

In this section, we further ablate on the \textit{number of layers} and \textit{hidden dimension size} of \texttt{NFPO}. The results are shown in \Cref{fig:as_more}.

For number of layers, \texttt{nfpo-2} achieves significant lower performance. This is because for odd-even-based masking we used, minimum 3 layers are required for expressiveness. For other number of layers (e.g., 4, 6 and 8) , they generally obtain similar performance. 

For hidden dimensions, they obtain quite similar results on simpler tasks like Go1JoystickRoughTerrain. For harder tasks like G1JoyStickRoughTerrain, \texttt{NFPO-512} obtains best performance while \texttt{NFPO-64} is the worst, and \texttt{NFPO-128} and \texttt{NFPO-256} obtain similar converged performance.

\subsection{\textcolor{black}{Additional tuning of \texttt{Meow}}}
\label{app:tune_meow}
\begin{figure}[htbp]
   \centering 
   \includegraphics[width=0.99\linewidth]{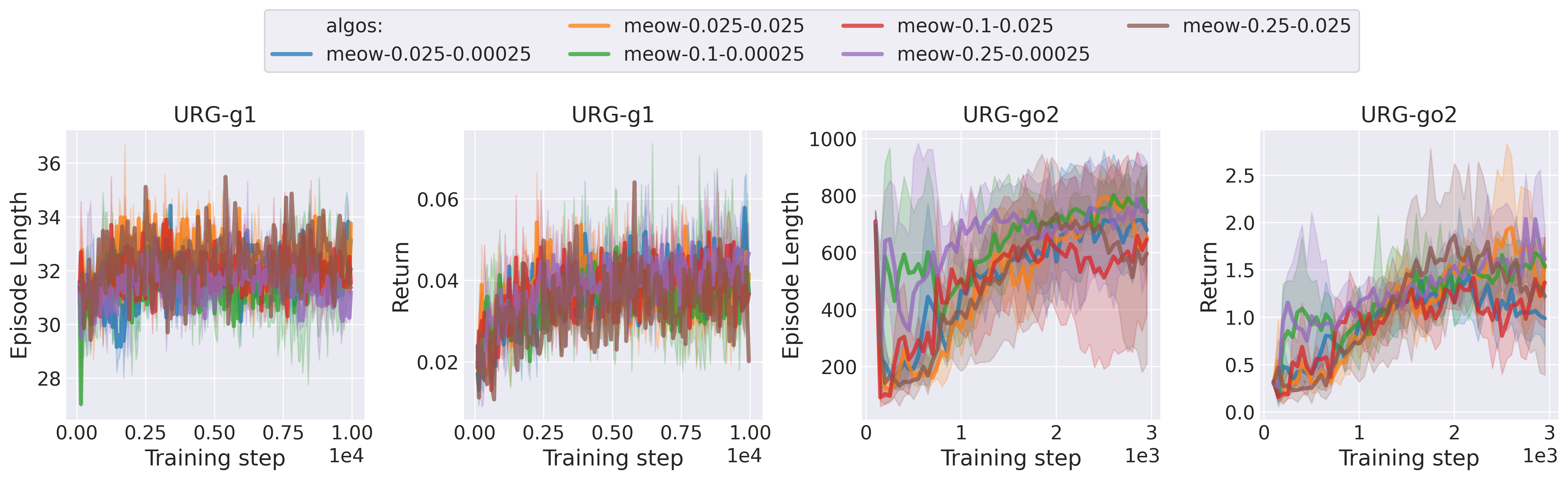}
   \caption{Performance of \texttt{Meow} over various tuned \texttt{polyak} and \texttt{alpha}. The experiments are in 3 seeds.}
   \label{fig:tune_meow}
\end{figure}

We additionally perform hyperparameters tuning for \texttt{Meow}. For the hyperparameters, we mainly tuned \texttt{polyak} and \texttt{entropy value} (\texttt{alpha}) similar to their original paper \citep{chaoMaximumEntropyReinforcement2024} and increased batch size to 10240 as recommended in \citet{seoFastTD3SimpleFast2025}. Other hyperparameters are kept the same as in \texttt{Meow}'s original codebase. The results are in \Cref{fig:tune_meow} and the legend is meow-\texttt{alpha}-\texttt{polyak} (the tuned \texttt{alpha} values are \{0.25, 0.1, 0.025\} and \texttt{polyak} values are \{2.5e-4, 2.5e-2\}). To facilitate the reproducibility, we also provide the source code we used for training \texttt{Meow} in Unitree RL Gym.

From the results, unfortunately, \texttt{Meow} generated no significant learning in these 2 environments. On URG-go2, its episode length is increased substantially, indicating it learns not to fall to ground to some extent but the episodic returns indicate that no meaningful locomotion or command following is learned. For URG-g1 which is more prone to fall, no significant learning in both episodic length and episodic return is observed.

\subsection{\textcolor{black}{Stabilized Deployment}}
\label{app:deploy}
Deployment on real-world whole-body-controlled humanoid robot is complex and could cause damages to the surrounding environments or assets. Hence, in our experiment, we supply the \textit{mode} of base Gaussian distribution to reduce the stochastics and help stabilize the deployment. This is similar to the common practice in \texttt{PPO} where people use the \textit{mean} ($\mu_\theta(s)$) as action in deployment (instead of sampling from $\mathcal{N}(\mu_\theta(s),\sigma_\theta(s))$ as in training).

This can also be seen as a similar case to what people do in  temperature-based generation where high temperature favors more random generation while low temperature favors more optimal generation. For \texttt{NFPO}, temperature could be controlled  in sampling $z \sim N(0, \tau I)$, for \texttt{PPO}, it's in $a \sim N(\mu_\theta(s), \tau \sigma_\theta(s))$. In our experiments, for both \texttt{PPO} and \texttt{NFPO}, if we sample actions with a larger temperature ($\tau$), both of them generate unstable behaviors. While using a low-temperature (a smaller $\tau$ that is not necessarily 0), both of them  generate stable behaviors.

\subsection{\textcolor{black}{The mode of Normalizing Flows}}
For RealNVP: 
\begin{align}
      f_{\theta_j}(a)_{d_j} &= a_{d_j} \\
   f_{\theta_j}(a)_{\backslash d_j} &= a_{\backslash d_j} \odot \exp( s_{\theta_j}(a_{d_j})) + t_{\theta_j}(a_{d_j}) \\
   \text{we have} \; \; \log \pi(a) &= \log q(f_\theta(a)) + \sum\nolimits_j \log \vert  \frac{df_{\theta_j}(a)}{da} \vert \\
   &= \log q(f_\theta(a)) + \sum_j s_{\theta_j}(a_{d_j})
\end{align}
As it uses non-constant scale ($s_\theta(a)$), the mode in prior distribution ($\mathcal{N}(0, I)$) may not be the mode of data distributions. However, the $\log q(f_\theta(a))$ term still favors smaller latent and in practice we find sampling near the mode of prior distribution is sufficient to generate stable behavior as in \Cref{app:deploy}. This is also similar to what \citet{kirichenko2020WhyNormalizingFlowsFailDetectOutofDistributionData} finds that RealNVP tends to use $t$ to reduce the effect of an increased $s$ to get a smaller latent, as a way to meet the 2 objectives simultaneously.

In \citet{chaoMaximumEntropyReinforcement2024}, as they adopt NICE \citep{dinhNICENonlinearIndependent2015} $\Bigl( f_{\theta_j}(a)_{\backslash d_j} = a_{\backslash d_j}  + t_{\theta_j}(a_{d_j}) \Bigr)$  which doesn't have varying scalar terms ($s_\theta(a) = 1$). Hence the mode in prior distribution is analytically the mode of data distribution.

\subsection{Pseudocode of NFPO}
\label{app:pseudo_code}
In this section, we provide a pseudo code of \texttt{NFPO} in \Cref{pseudo_code}.

\begin{algorithm}[htbp]
  \DontPrintSemicolon
  \SetAlgoLined
    \newcommand{\nonl}{\renewcommand{\nl}{\let\nl\oldnl}}
  \caption{Pseudo code of NFPO}
    \label{pseudo_code}
  \SetKwInput{PaFor}{Parallel For}
  \KwIn{parallel environment $e$, initial network parameter $\theta$, clip ratio $\epsilon$}

  \SetKwProg{Fn}{Function}{}{end}
  \SetKwFor{ParallelFor}{Parallel for}{do}{end}

  \While{training}{
    $B \gets []$

   \For{$t = 1$ \KwTo $T$}{
      $z \sim q(z)$

      $a \gets f_\theta^{-1}(z; o)$ \;
      $\log a = \log(f_\theta(a)) + \sum\nolimits_j \log \vert \frac{df_{\theta_j}(a; o)}{da} \vert$ \;
      $o^\prime$, $r$, $d$ $\gets$ \FuncSty{step}$(e, a)$

      $B \gets $ \FuncSty{append}$(B, (o, a, r, d, \log a, o^\prime))$

   }

   $\tilde{A} \gets $ \FuncSty{GAE}$(B)$  \tcp{estimate advantage via GAE}

   \For{$n = 1$ \KwTo $N$} {

      $bs \gets$ \FuncSty{partition}$(B)$ \tcp{partition large buffer $B$ to smaller $bs$}

      \For{$m = 1$ \KwTo $M$} {

         $b \gets bs[m] $

         $agent \gets $ \FuncSty{update}$(agent, b)$
      }
   }

  }
    \Fn{\FuncSty{update(agent,b)}}{

      \tcc{the update of value function is not changed and omitted}

      $o, \pi_{\text{old}}(a) \gets b$

      $r(\theta) \gets$  $\frac{\pi_\theta(a; o)}{\pi_{\text{old}}(a)}$ \;

      $L_\theta \gets$ $\nabla_\theta \frac{1}{\vert b\vert}\sum_b \Bigl[\min \bigl( r(\theta)\tilde{A}(s,a), \text{clip} (r(\theta), 1-\epsilon, 1+\epsilon) \tilde{A}(s,a) \bigr)\Bigr]$

      $agent \gets$ \FuncSty{SGD}$(agent, L_\theta)$

    \KwRet{$agent$} \;

  }

\end{algorithm}

\subsection{Details of hyperparameters}
We list the details of training hyperparameters in \Cref{app:tbl_hyper}. For \texttt{PPO} and \texttt{NFPO}, most components and hyperparameters could be shared  and major difference lies in the actor network parameterization which we have aligned in terms of parameter numbers. 

\begin{table}[htbp]
    \centering
      \caption{Hyperparameter settings}
      \label{app:tbl_hyper}
   \begin{threeparttable}
      \newcolumntype{C}[1]{>{\centering\arraybackslash}m{#1}}

        \begin{NiceTabular}{|C{2.0cm}|C{2.5cm}|C{3.3cm}|C{2.8cm}|}[hvlines]
            \CodeBefore
                \rowcolor{lightgray}{1-1}
            \Body
           \Block{1-2}{Hyperparameter} &  & Value & Remarks \\
    \Block{9-1}{Shared}    &   GAE $\lambda$ & 0.95 & \\
        &   discount $\gamma$ & 0.99 & \\
        &   value loss coefficient & 1.0 & \\
        &   grad clip & 1.0 & \\
        &   learning epoch & 5 & \\
        &   learning minibath & 4 & \\
        &   entropy loss coefficient & $10^{-3}$ & only for PPO \\
        &   step length & 24 & \\
        &   desired KL divergence & $10^{-2}$ & only for PPO adaptive \\
  \Block{5-1}{Unitree RL Gym}      &   learning rate & $10^{-3}$ & \\
        &   value network hidden dims & $[32]$ & 1 layer network with hidden dim 32 \\ 
        &   PPO actor network hidden dimes & $[96, 96, 64]$ &  \\
        &   NFPO actor network hidden dimes & $4 \times [64]$ & 4 layers, each layer with 64 hidden dim \\
        &   number of environments & 4096 & \\
 \Block{5-1}{Mujoco Playground}       &   learning rate & $5e^{-4}$ & $3e^{-4}$ for manipulation tasks\\ 
 &   value network hidden dims & $[512, 256, 128]$\\ 
        &   PPO actor network hidden dimes & $[512, 256, 128]$ &  \\
        &   NFPO actor network hidden dimes & $4 \times [256]$ & \\
        &   number of environments & 2048 & \\
            
        \end{NiceTabular}
    \end{threeparttable}
\end{table}

\subsection{Details of Hardware and Computation Efficiency}
\label{app:runtime}

Most of our experiments run on a 16 CPU x 1 Nvidia A100-40G GPU except for \texttt{Meow} which needs more than 40G GPU memory and run on A100-80G.

Depending on specific training settings, the wall clock time of each experiment varies and we provide the wall clock time of \texttt{PPO} and \texttt{NFPO} on Unitree RL Gym's g1 environment in \Cref{app:tbl_runtime}:

\begin{table}[htbp]
    \centering
      \caption{Runtime of \texttt{PPO} and \texttt{NFPO} on Unitree Gym's g1. Calculated using 10 seeds.}
      \label{app:tbl_runtime}
   \begin{threeparttable}
      \newcolumntype{C}[1]{>{\centering\arraybackslash}m{#1}}

        \begin{NiceTabular}{|C{2.0cm}|C{4.5cm}|C{3cm}|C{1.3cm}}[hvlines]
            \CodeBefore
                \rowcolor{lightgray}{1-1}
            \Body
           Algorithm  & Runtime (s) & Runtime (h) & Ratio \\
           \texttt{PPO} & 13500.94 $\pm$ 628.29 &3.75 $\pm$ 0.17 & 1 \\
           \texttt{NFPO} & 16036.06 $\pm$ 795.74 & 4.45 $\pm$ 0.22 & 1.19\\
            
        \end{NiceTabular}
    \end{threeparttable}
\end{table}

\pagebreak
\pagebreak

\end{document}